\documentclass[twocolumn,letterpaper]{IEEEtran}
\usepackage[flushleft]{threeparttable}
\usepackage{cite}
\usepackage{times}
\usepackage{amsmath}
\usepackage{bm}
\usepackage{mathtools}
\usepackage{amsmath}
\usepackage{amsthm}
\usepackage{amssymb}
\usepackage{subfigure}
\usepackage{verbatim}
\usepackage{graphicx}
\usepackage[thinlines,thiklines]{easybmat}
\usepackage{latexsym}
\usepackage[dvipsnames]{xcolor}
\usepackage{cite}
\usepackage{stmaryrd}
\usepackage{cuted}
\usepackage{lipsum}
\usepackage{float}
\usepackage{multirow}
\usepackage{textcomp}
\usepackage[linesnumbered,ruled,vlined]{algorithm2e}
\usepackage{bookmark}

\newsavebox{\twosubbox}

\def\sign{\mathop{\rm sign}\nolimits}
\def\s{\mathop{\rm s}\nolimits}
\def\c{\mathop{\rm c}\nolimits}
\newcommand{\bs}[1]{\ensuremath{{\boldsymbol{#1}}}}

\def\diag{\mathop{\rm diag}\nolimits}

\def\diag{\mathop{\rm diag}\nolimits}
\def\sign{\mathop{\rm sgn}\nolimits}

\makeatletter
\def\ALG@special@indent{%
	\ifdim\ALG@thistlm=0pt\relax
	\hskip-\leftmargin
	\else
	\hskip\ALG@thistlm
	\fi
}
\newcommand{\Notations}[1]{\item[]\noindent\ALG@special@indent \textbf{Notations:}\ #1}

\makeatletter
\newcommand{\multiline}[1]{%
	\begin{tabularx}{\dimexpr\linewidth-\ALG@thistlm}[t]{@{}X@{}}
		#1
	\end{tabularx}
}
\makeatother

\DeclareFontFamily{OMX}{yhex}{}
\DeclareFontShape{OMX}{yhex}{m}{n}{<->yhcmex10}{}
\DeclareSymbolFont{yhlargesymbols}{OMX}{yhex}{m}{n}
\DeclareMathAccent{\wideparen}{\mathord}{yhlargesymbols}{"F3}

\title{Autonomous Bikebot Control for Crossing Obstacles with Assistive Leg Impulsive Actuation~\thanks{This work was supported  in part by the US NSF award CNS-1932370 (Yi) and NSFC awards 52175033 and U21A20120 (Liu). ({\em Corresponding authors: Jingang Yi} and {\em Tao Liu}.)}}

\author{Feng Han\thanks{First two authors equally contributed to this work.}, Xinyan Huang\thanks{X. Huang, Z. Wang, and T. Liu are with the State Key Lab of Fluid Power and Mechatronic Systems and the School of Mechanical Engineering, Zhejiang University, Hangzhou, Zhejiang 310027 China (email: hxinyan@zju.edu.cn; zenghao\_wang@zju.edu.cn; {liutao@zju.edu.cn}).}, Zenghao Wang, Jingang Yi\thanks{F. Han and J. Yi are with the Department of Mechanical and Aerospace Engineering, Rutgers University, Piscataway, NJ 08854 USA (e-mail: fh233@scarletmail.rutgers.edu; jgyi@rutgers.edu).}, and Tao Liu}

\begin{document}
\maketitle

\begin{abstract}
As a single-track mobile platform, bikebot (i.e., bicycle-based robot) has attractive navigation capability to pass through narrow, off-road terrain with high-speed and high-energy efficiency. However, running crossing step-like obstacles creates challenges for intrinsically unstable, underactuated bikebots. This paper presents a novel autonomous bikebot control with  assistive leg actuation to navigate crossing obstacles. The proposed design integrates the external/internal convertible-based control with leg-assisted impulse control. The leg-terrain interaction generates assistive impulsive torques to help maintain the navigation and balance capability when running across obstacles. The control performance is analyzed and guaranteed. The experimental results confirm that under the control design, the bikebot can smoothly run crossing multiple step-like obstacles with height more than one third of the wheel radius. The comparison results demonstrate the superior performance than those under only the velocity and steering control without leg assistive impulsive actuation.
\end{abstract}


\section{Introduction}
\label{Sec_Introduction}

As a single-track mobile platform, bikebot (i.e., bicycle-based robot) offers attractive navigation capability such as high-speed and high-energy efficiency on off-road terrains~\cite{blueteam2,SEEKHAO2020102386Development}. Comparing with four-wheel, double-track platform, bikebot has high agility but potentially suffers intrinsic unstable, underactuation disadvantages. One major control challenge of underactuated bikebot lies in the non-minimum phase, unstable roll dynamics~\cite{HanRAL2021}. Modeling and control of single-track bikebots attracted extensive attention in past decades~\cite{Astrom,GetzPhD,YiICRA2006} and various autonomous bikebots have been developed and demonstrated~\cite{Tanaka2009,HeROBIO2015,chen2021robust}. An external and internal convertible (EIC) form was presented for bikebot dynamics~\cite{GetzPhD} and the EIC-based control has been successfully demonstrated~\cite{YiICRA2006,ZhangPhD2014,WangICRA2017}. Besides steering and velocity control, additional assistive devices, such as reaction wheels~\cite{chen2021robust,CuiIROS2020}, control moment gyroscope~\cite{ZhangICRA2014,kim2015stable} and inverted pendulum~\cite{SEEKHAO2020102386Development}, are usually installed on the bikebot to enhance balance and tracking performance.

\begin{figure*}
    \centering
	\subfigure[]{
	\label{Fig_BikePhoto}
	\includegraphics[width=2.8in]{Figure/Fig_BikePhoto3.pdf}}
    \hspace{-0mm}
	\subfigure[]{
	\label{Fig_Impact_3D}
	\includegraphics[height=1.85in]{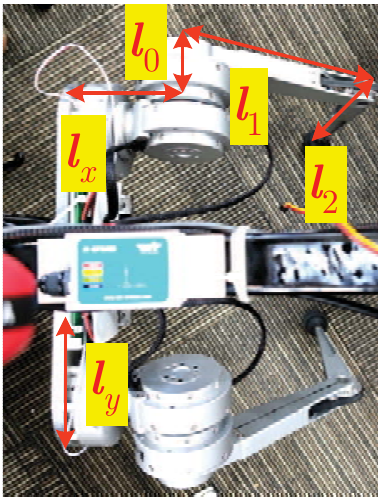}}
    \hspace{-0mm}
	\subfigure[]{
	\label{Fig_Impact_2D}
	\includegraphics[height=1.85in]{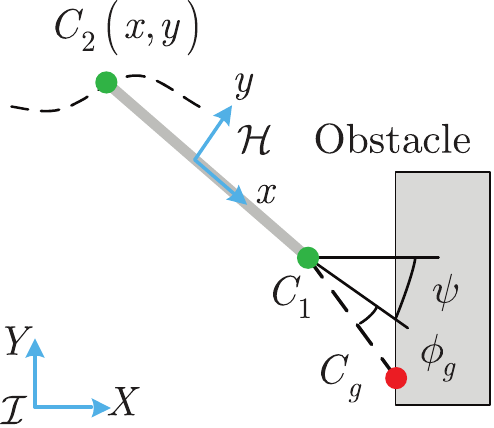}}
	\caption{(a) The prototype of the bikebot system with assistive legs. (b) Schematic of the systems modeling. (c) Schematic of the wheel-obstacle impact.}
	\vspace{-3mm}
\end{figure*}

All of the above studies consider that the autonomous bikebots navigate on flat solid surfaces or smoothly curving terrains. For outdoor off-road applications, bikebots often need to run crossing step-like obstacles such as curbs, steps, etc. The major challenges for bikebots to navigate across these terrains include complex wheel-obstacle interactions and limited stabilization actuation~\cite{SEEKHAO2020102386Development}. It is well-known that legged locomotion has advantages to navigate on uneven terrains~\cite{shkolnik2011bounding,valsecchi2020quadrupedal} and using both wheel and leg modalities enables mobile robots to adapt to different operation environments and expand their workspace~\cite{Peng2020Coordinated}. In~\cite{Chen2017TurboQuad}, a novel leg–wheel transformable robot was designed to perform motions in wheeled, legged trotting and walking modes. Inspired by this observation, we propose a leg-assisted impulse control on bikebot to help balance and run across step-like obstacles.

When running across an obstacle, the wheel-obstacle impact disturbs the bikebot stability and the steering and velocity control only provides limited actuation to maintain balance~\cite{kant2018impulsive}. In this case, assistive legs can provide additional balance torques. Due to high moving speed, the leg-terrain contact should be kept as short as possible to minimize its negative impact on bikebot motion. Therefore, a leg-terrain impulsive interaction is desirable for bikebot balance assistance. Impulse control has been used to regain the stability of the dynamical system and enlarge the region of attraction in robotic applications~\cite{kant2019estimation}. In~\cite{weibel1996control}, an impulse control algorithm was developed for a single degree-of-freedom (DOF) Hamiltonian system such as the cart-pendulum example. In~\cite{kant2018impulsive}, the optimal control strategy with finite numbers of the impulse inputs was designed to swing up an inertia-wheel pendulum. One main difference in this work is the integration of the EIC-based control with the impulse control for underactuated balance robots such as bikebot.

In this paper, we present a novel control design of an autonomous bikebot for running across obstacles with two assistive legs. A dynamic model of the bikebot system is first presented. The wheel-obstacle impact is modeled and its effects on bikebot dynamics are estimated. A machine learning-based method is used to enhance the estimated post-wheel-obstacle impact motion through the onboard inertial measurement unit (IMU) measurements. The impulsive torque control is designed through an optimization-based method for real-time implementation. Integration and performance analysis of the impulse control and the EIC-based design are discussed. We demonstrate the design and control performance with extensive experiments. Compared with other assistive devices (e.g., reaction wheels, gyroscopic balancers) that often suffer limited actuation, the leg impulsive torque design provides continuous assistive balance torques and also potentially serves as an active recovery and support mechanism. The main contributions of the work are twofold. First, this work considers the use of assistive leg and impulse control for wheeled robot to navigate crossing obstacles. To the authors' best knowledge, few research work explicitly discusses the mechatronic design for helping unstable mobile robots crossing obstacles. The integrated wheel-leg locomotion control is also novel and can be further extended for enabling new bikebot applications. Second, the impulse control is designed though an optimization approach and it has advantage to integrate with the previously developed simultaneously trajectory tracking and balance control for underactuated mobile robots.

The remainder of this paper is outlined as follows. In Section~\ref{Sec_Problem_Formulation}, we present the dynamic model and control of the bikebot system. Section~\ref{Sec_Wheel_Obstacle_Impact} discusses the wheel-obstacle impact and presents the machine learning-based impact estimation enhancement. The impulse control is discussed in Section~\ref{Sec_Impulsive_Control} and experiments are presented in Section~\ref{Sec_Experiment}. Finally, Section~\ref{Sec_Conclusion} summarizes the concluding remarks.

\section{Problem Statement and Systems Dynamics}
\label{Sec_Problem_Formulation}

\subsection{Problem Statement}

Fig.~\ref{Fig_BikePhoto} shows a prototype of the autonomous bikebot with two assistive legs. The bikebot is driven by a wheel hub motor and the steering actuation is also driven by a DC motor. Two 3-DOF assistive legs are installed at the bikebot's lower frame with one leg on each side symmetrically. As shown in the figure, when moving across a step-like obstacle (i.e., a wooden board), the wheel-obstacle impact generates a large disturbance for the bikebot motion and the platform can lose stability and control. The problem considered is to design the leg-assisted impulsive torque control together with the steering and velocity control to maintain the balance and trajectory tracking when running across obstacles.

\subsection{System Dynamics and EIC-based Control}

Fig.~\ref{Fig_Impact_3D} illustrates the schematic of the systems configuration. The bikebot roll, yaw, and steering angles are denoted as $\psi$, $\varphi_b$, and $\phi$, respectively. Three frames are used in systems modeling: an inertial frame $\mathcal{I}$, body frame $\mathcal{B}$, and moving frame $\mathcal{H}$. Frame $\mathcal{I}$ is fixed on the ground with the $Z$-axis perpendicular to the ground, while the origin of $\mathcal{H}$ is located at the middle point of line $C_1C_2$, with the $z$-axis pointing upwards and $x$-axis aligned with $C_1C_2$, where $C_1$ and $C_2$ are the front and rear wheel contact points, respectively. Frame $\mathcal{B}$ is obtained from $\mathcal{H}$ by rotating $\varphi_b$ around the $x$-axis with the same origin. The center of mass of the bikebot is denoted as $G$ and its coordinate in $\mathcal{B}$ is $[l_G\;0\;h_G]^T$.

\subsubsection{Bikebot Model and EIC-based Control}
\label{control0}

The bikebot planar motion is captured by the $C_2$ position vector $\bs{r}=[x \; y]^T$ in $\mathcal{I}$. The nonholonomic constraint at $C_2$ gives $\dot{x}=v\c_\psi, \; \dot{y}=v \s_\psi$, where $v$ is the bikebot velocity magnitude and notations $\c_\psi:=\cos \psi$ and $\s_\psi:=\sin \psi$ are used for $\psi$ and other angles. Taking the third order derivative of $\bm r$ produces
\begin{equation}
\label{Eq_Bikebot_Kine}
\bs{r}^{(3)}=\begin{bmatrix}
x^{(3)} \\
y^{(3)}
\end{bmatrix}
=-\underbrace{\begin{bmatrix}
v \dot{\psi} \c_{\psi}+2 \dot{v} \s_{\psi} \\
v \dot{\psi} \s_{\psi}-2 \dot{v} \c_{\psi}
\end{bmatrix}}_{\bm R_{\psi}} \dot{\psi}
+\underbrace{\begin{bmatrix}
\c_{\psi} & -v \s_{\psi} \\
\s_{\psi} & v \c_{\psi}
\end{bmatrix}}_{\bm K_{\psi}} \bs{u},
\end{equation}
where $\bs{u}=[u_v \; u_\psi]^T$, $u_v=\ddot{v}$, and $u_\psi=\ddot{\psi}$. From~\cite{WangTASE2019}, the yaw angular rate is calculated as $\dot{\psi}=\frac{v\c_\varepsilon}{l \c_{\varphi_b}} \tan \phi$ and $u_\psi$ is
\begin{equation}\label{Eq_U_psi}
    u_\psi = \frac{{{{\dot v}}\tan \phi {\c_\varepsilon }}}{{l{\c_{\varphi_b} }}} + \frac{v\c_\varepsilon}{l\c_{\varphi_b}}( {\sec _\phi ^2\dot \phi  + \tan \phi \tan \varphi_b \dot \varphi_b }),
\end{equation}
where $l$ is the wheelbase and $\varepsilon$ is the steering caster angle. The roll motion of the bikebot is captured as~\cite{WangTASE2019}
\begin{equation}
\label{Eq_Bikebot_Roll_2}
  J_{t} \ddot{\varphi}_b=f_1(\varphi_b)+h(\varphi_b) u_{\psi}
\end{equation}
where $f_1(\varphi_b)=m_b{h_G}\dot \psi {\c_{{\varphi _b}}}v + m_bh_G^2{\dot \psi ^2}{\s_{{\varphi _b}}}{\c_{{\varphi _b}}} + mgh_G{\s_{{\varphi _b}}}$, $h(\varphi_b)=m_bh(l_G+l/2){\c_{{\varphi _b}}}$,  $J_{t}=J_b+m_b h_G^2$, $J_b$ and $m_b$ are the mass moment of inertia around the $x$-axis and mass of the bikebot, respectively. Substituting~(\ref{Eq_U_psi}) into~(\ref{Eq_Bikebot_Roll_2}) and collecting terms in $\tan\phi$, we obtain the steering-induced balance torque
\begin{equation}
\label{Eq_Steering_Torque}
\tau_s = \frac{m_bh_G\c_\varepsilon v}{l}\left[\frac{(l_G+l/2)\dot{\varphi}_b}{l}\tan \varphi_b + \frac{l_G\dot v}{v}-v \right]\tan \phi.
\end{equation}
Using $\tau_s$, the roll motion model~(\ref{Eq_Bikebot_Roll_2}) becomes
\begin{equation}
\label{Eq_Bikebot_Roll_3}
  {J_t}{{\ddot \varphi }_b} = f_2({\varphi _b})+\tau_s
\end{equation}
where  $f_2({\varphi _b})=m g h_{G} \s_{\varphi_{b}}+m_{b} h_{G}^{2} \frac{v^{2} \mathrm{c}_{\varepsilon}^{2}}{l^{2}} \tan \varphi_{b} \tan ^{2} \phi+m_{b} h_G (l_{G}+l/2) \frac{v \mathrm{c}_{\varepsilon}}{l} \sec _{\phi}^{2} \dot{\phi}$.

For a given trajectory $\bm r_d=[x_d \;y_d]^T$, the tracking control
\begin{equation}
\label{Eq_Trakcing_Input}
    \bm u= \bm K_\psi ^{ - 1}( {{\bm R_\psi } + {\bm u_r}} ), \bm u_r= \bm r_d^{(3)}+a_2\ddot{\bm e}_r+a_1\dot{\bm e}_r+a_0\bm e_r
\end{equation}
drives the tracking error $\bm e_r=\bm r- \bm r_d$ to converge to zero asymptotically, where $a_i>0$, $i=0,1,2$. To incorporate the balance task, the roll motion is enforced to move along the balance equilibrium manifold under $u_{\psi}$ in~(\ref{Eq_Trakcing_Input}) as $\mathcal{E}= \left\{ \varphi _b^e:\, f_1(\varphi_b^e)+h(\varphi_b^e) u_{\psi}=0 \right\}$. The balance control is updated as
\begin{equation}
\label{Eq_Updated_Steering_Control}
    \bar u_\psi= h^{-1}(\varphi_b)\left[J_t u_b - f_1(\varphi_b)\right], \, u_b= \ddot \varphi _b^e+b_1 e_b+b_0 e_b,
\end{equation}
where $b_0, b_1>0$ and $e_b=\varphi_b-\varphi_b^e$. The final control design incorporates both the above trajectory tracking and balance control, that is,
\begin{equation}
\label{Eq_EIC_Control}
{\bm u}=[u_v \; \bar u_\psi]^T.
\end{equation}
Under~(\ref{Eq_EIC_Control}), the closed-loop system errors for both the trajectory tracking and roll balance converge to a small ball around zero exponentially~\cite{GetzPhD}.

\subsubsection{Assistive Leg Kinematics}

The two 3-DOF legs are attached to the frame at $S$ on the bikebot; see Fig.~\ref{Fig_Impact_3D}. The position of $S$ in $\mathcal{B}$ is $\bs{r}_S=[l_b\; 0\; h_b]^T$. The $j$th joint angle and link length are denoted as $\theta_{i,j}$ and $l_{i,j}$, $j=0,1,2$, $i=L,R$, for left and right legs, respectively. The position of the first joint relative to the $S$ is $[l_x \;l_y \;0]^T$. We model the leg kinematics in $\mathcal{H}$. For left and right legs, the homogeneous transformation matrices are then
\begin{equation}
    \mathcal{A}_{i}^\mathcal{H} = \mathcal{A}_\mathcal{B}^\mathcal{H}\left( {{\varphi _b}} \right)\mathcal{A}_{i}^\mathcal{B}\left(\bs{\theta}_i\right), \; i=L,R,
\end{equation}
where $\bs{\theta}_i=[\theta_{i0}\; \theta_{i1}\;\theta_{i2}]^T$, $\mathcal{A}_\mathcal{B}^\mathcal{H} \in \mathbb{R}^{4\times4}$ is the homogeneous transformation matrix from $\mathcal B$ to $\mathcal H$, and $\mathcal{A}_{i}^\mathcal{B}$ describes the leg foot pose in $\mathcal{B}$. For the left leg, the position vector is $\bm r_L=[r_x \; r_y \; r_z]^T$ in $\mathcal H$ with
\begin{subequations}\label{Eq_Leg_Kinamatics}
\begin{align}
    r_x& = l_x - l_b + l_{2}\s_{\theta_{1}+\theta_{2}} + l_{1}\s_{\theta_{1}},\\
    r_y& = \c_{\varphi_b}(l_y + l_{0}\c_{\theta_{0}}- l_{1}\s_{\theta_{0}}\c_{\theta_{1}} - l_{2}\s_{\theta_{0}}\c_{\theta_{1} + \theta_{2}})\nonumber\\
    &\quad-\s_{\varphi_b}(h_b + l_{0} \s_{\theta_{0}} + l_{1} \c_{\theta_{0}} \c_{\theta_{1}} + l_{2} \c_{\theta_{0}} \c_{\theta_{1} + \theta_{2}}),\\
    r_z&= -\c_{\varphi_b}(h_b + l_{0} \s_{\theta_{0}} + l_{1} \c_{\theta_{0}} \c_{\theta_{1}} + l_{2}\c_{\theta_{0}}\c_{\theta_{1} + \theta_{2}})\nonumber \\
    &\quad-\s_{\varphi_b}(l_y + l_{\theta_{0}}\c_{\theta_{0}}-l_{1}\c_{\theta_{1}}\s_{\theta_{0}}-l_{2}\s_{\theta_{0}}\c_{\theta_{1} + \theta_{2}}).
\end{align}
\end{subequations}
Similar kinematics are obtained for the right leg.

\section{Wheel-Obstacle Impact Estimation}
\label{Sec_Wheel_Obstacle_Impact}

In this section, we present the wheel-obstacle impact model and then discuss a machine learning-based method to enhance the impact estimation accuracy.

\subsection{Wheel-Obstacle Impact Model}

Fig.~\ref{Fig_Impact_2D} illustrates the wheel-obstacle schematic. The front wheel is considered to hit the obstacle at point $C$. The wheel-obstacle interaction is considered as an elastic impact. The step-like obstacle has height $h_o$ and the impact angle is denoted as $\phi_c=\phi_g+\psi$, where $\phi_g=\tan^{-1}\left(\frac{\c_\varepsilon}{\c_{\varphi_b}}\tan\phi\right)$ is the projected steering angle (along the front wheel plane). Denoting coordinate $\bm q=[x \;y \;z \;\psi \; \varphi_b]^T$ (i.e., position of $C_2$ and yaw and roll angles), the impact model is written as
\begin{equation}
\label{Eq_Impact}
\bm D (\dot{\bs{q}}_f^+- \dot{\bs{q}}_f^-) = \bs{I}_\mathbb{I}, \; \bm D = \begin{bmatrix} m_b\bm{I}_3 & \bs{0} \\ \bs{0} & \bm{J}_b\end{bmatrix},
\end{equation}
where $\bm{I}_3$ is a $3\times 3$ identity matrix, $\bs{J}_b=\diag(J_z,J_t)$, and $J_z$ is the moment of inertia matrix of the bikebot around $z$-axis at $G$. We use superscripts ``$+$'' and ``$-$'' to represent the variables immediately after and before the impact (or applied impulse) and subscript ``f'' indicates the wheel-obstacle impact. The impulsive torque $\bs{I}_\mathbb{I}$ in~(\ref{Eq_Impact}) is related with the impact force ${\bm f}_\mathbb{I}$ at $C$ as
\begin{equation}
\label{Eq_Impact_Mapping}
\bs{I}_\mathbb{I} =\bm {J}^T_C{\bm f}_\mathbb{I},
\end{equation}
where $\bm{J}_C=\frac{\partial \bm r_C( \bm q)}{\partial \bm q}  \in \mathbb{R}^{3\times 5}$ is the Jacobian matrix,
\begin{equation}
	\bm{r}_C(\bm q) = [ x + l\c_\psi  + L \c_{\psi_c}\;
		y + l\s_\psi  + L \s_{\psi_c}\; z + h_o]^T ,
\end{equation}
$L = \sqrt{2R_wh_o-h_o^2}$, and $R_w$ is the wheel radius.

Since $\bs{I}_\mathbb{I}$ in~(\ref{Eq_Impact}) is unknown, additional equations are needed to obtain $\dot{\bm q}_f^+$. Given an elastic impact at $t=t_f$, we denote $\dot{\bm r}_C|_{t=t_f^+}=\bm{J}_C\dot{\bm q}_f^+=\bm \epsilon$, where $\bm \epsilon$ is the elastic velocity after impact and obtained through the experiment data. Combining~(\ref{Eq_Impact}) and ~(\ref{Eq_Impact_Mapping}), we obtain
\begin{equation}\label{Eq_Impact_Dynamics}
\underbrace{\begin{bmatrix}  \bm D &- \bm J^T_C \\ \bm J_C& \bf 0 \end{bmatrix}}_{\bs{D}_i}\begin{bmatrix}
 \dot {\bm q}_f^+ \\ \bm f_\mathbb{I} \end{bmatrix} = \begin{bmatrix} \bm D\dot{\bm q}_f^- \\  \bm \epsilon\end{bmatrix}.
\end{equation}
Because of full rank for $\bs{D}_i$, we solve and obtain $\dot{\bm q}_f^+$ by inverting~(\ref{Eq_Impact_Dynamics}).

\subsection{LSTM Based Impact Model}

To account for model uncertainties and modeling errors, we extend~(\ref{Eq_Impact_Dynamics}) to use a machine learning method to improve the estimation results. We consider the enhanced post-impact estimation as
\begin{equation}\label{Eq_LSTM_Impact}
    \dot {\bm q}_f^*  = \dot {\bm q}_f^+  + \delta \dot {\bm q}^ +
\end{equation}
where $\dot {\bm q}_f^*$ is the enhanced velocity after impact, $\dot {\bm q}_f^+$ is obtained from~(\ref{Eq_Impact_Dynamics}), and $\delta \dot {\bm q}^ +$ denotes the correction term.

\begin{figure}[h!]
	\centering
	\includegraphics[width=3.2in]{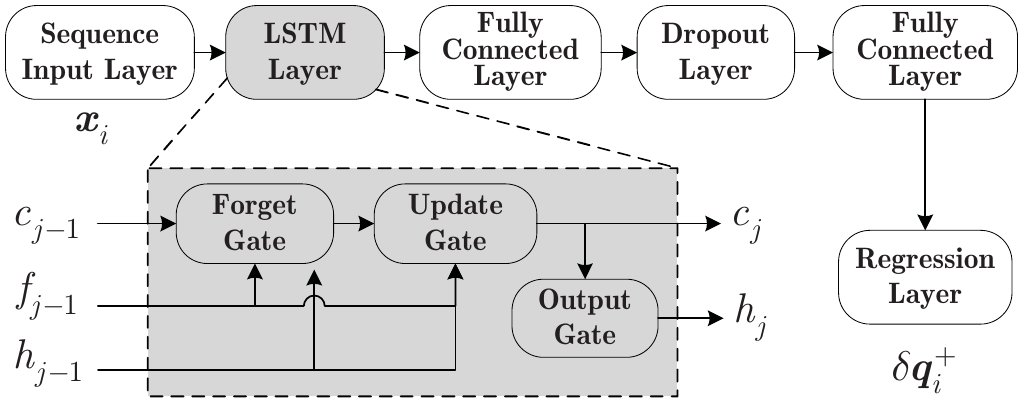}
	\vspace{-1mm}
	\caption{Schematics of the LSTM-based estimation enhancement scheme.}
	\label{Fig_LSTM_Diagram}
	\vspace{-2mm}
\end{figure}

We use a long short time memory (LSTM) neural network model to estimate $\delta \dot {\bm q}^ +$. Fig.~\ref{Fig_LSTM_Diagram} illustrates the structures of the LSTM network, including the input layer, LSTM layers, fully connected layers and regression layer. The neuron numbers at each layer are different and we use multiple layers to build a deep neural network. The network builds the relationship between the bikebot motion measurements and the velocity correction term $\delta \dot {\bm q}^+$. The LSTM network model is considered primarily because of its properties to capture nonlinear temporal relationship between input and output and its capability to be implemented for real-time applications.

The LSTM network is trained offline and prediction for $\delta \dot {\bm q}^+$ is used in real-time impulsive control. The training data is denoted as $\mathbb{D}=(\left\{\bm x_i \right\}_{i=1}^M,\left\{\delta \dot {\bm q}^+_i  \right\}_{i=1}^M)$, where $M$ is the number of data points, $\bm x_i=\{\bs{a}_{j}, \bs{\omega}_j, v_j, \varphi_{bj},\phi_j, \psi_j \}_{j=i-N+1}^{i}$ contains $N$ observations of the IMU's accelerometer and gyroscope measurements (i.e., $\bs{a}_j \in \mathbb{R}^3$ and $\bs{\omega}_j \in \mathbb{R}^3$, respectively) and the bikebot motion and steering actuation (i.e., $v, \varphi_b, \phi, \psi$). The training data $\delta \dot {\bm q}^+_i $ is obtained from the difference of bikebot motion from the motion capture system and the impact model~(\ref{Eq_Impact_Dynamics}). The LSTM network is to capture the relationship between $\bm x_i$ and $\delta \dot {\bm q}^+_i$. In online prediction phase, the trained LSTM network is stored on the local computer and the variable $\bm x$ at the impact is fed into the trained LSTM network to predict $\delta \dot {\bm q}^ +$ to enhance the analytical model estimation. We will show the experimental validations in Section~\ref{Sec_Experiment}.

\section{Assistive Impulsive Control Design}
\label{Sec_Impulsive_Control}

In this section, we first present the assistive impulse control and then discuss the integration with the EIC-based control.

\vspace{-2mm}
\subsection{Impulse Control Design}
\label{Sec_Impulse_Control_Design}

Denoting the interaction force as $\bs{F}$, the applied torque on the bikebot due to leg-terrain interaction is $\bs\delta \tau = \bm r_i \times \bs{F}$, where $\bm r_i$, $i=L,R$, is the leg contact point position from~(\ref{Eq_Leg_Kinamatics}). Given the left (right) leg joint torque vector $\bs{\tau}_\theta \in \mathbb{R}^3$, force $\bs{F}$ is calculated by $\bs{F}=(\bm{J}^T_\theta\left(\bm R_\mathcal{H}^\mathcal{B}\right)^T)^{-1}\bm \tau_\theta$, where $\bm R_\mathcal{B}^\mathcal{H}$ is the rotation matrix from $\mathcal{B}$ to $\mathcal{H}$ and $\bm{J}_\theta$ is the Jacobian matrix of the foot. Therefore, we obtain the applied torque
\begin{equation}
\bs\delta \tau = \bm r_i \times \bs{F}=\bm r_i \times (\bm{J}^T_\theta\left(\bm R_\mathcal{H}^\mathcal{B}\right)^T)^{-1}\bm \tau _\theta.
\label{Eq_Leg_Impule_Torque}
\end{equation}
The impulsive balance torque is the component of $\bs\delta_\tau$ along the roll direction, i.e., $\delta\tau_x=\bm \delta \tau \cdot {\bm e}_x$, $\bm e_x=[1\;0\;0]^T$. For balance purpose, we mainly focus on the control design of $\delta\tau_x$ and the torque components in the $y$-axis ($\delta \tau_y$) and $z$-axis ($\delta \tau_z$) directions are undesirable and therefore should be minimized.

We consider that impulsive torque $\bs{\delta}\tau$ is applied at $t=t_\tau$ with a short duration $\kappa$, that is, from $t_\tau$ to $t_\tau^+:=t_\tau+\kappa$. Under $\delta\tau_x$ the bikebot roll velocity experiences a discontinuous jump, while the roll angle remains the same. With $\delta \tau_x$, using~(\ref{Eq_Bikebot_Roll_2}) we obtain
\begin{equation*}
\mathop {\lim }_{\kappa  \to 0} \int_{t_\tau}^{t_\tau^+} J_t\ddot \varphi_bdt  = \mathop {\lim }_{\kappa \to 0} \int_{t_\tau}^{t_\tau^+} [f_1(\varphi_b) + h(\varphi_b)u_\psi + \delta \tau_x] dt.
\end{equation*}
Note that $\mathop {\lim}_{\kappa  \to 0} \int_{t_\tau}^{t_\tau^+} [f_1(\varphi_b) + h(\varphi_b)u_\psi] dt =0$ since the roll angle remains unchanged during $[t_\tau,t_\tau^+)$ for an ideal impulse. Therefore, from the above equation we obtain
\begin{equation}
\label{Eq_Impulse_Effect}
  J_t \Delta \dot \varphi_b = \tau_\mathbb{I}:=\int_{t_\tau}^{t_\tau^+} \delta\tau_x dt,
\end{equation}
where $\Delta \dot \varphi_b = \dot \varphi_{b\tau}^+ -\dot \varphi_{b\tau}^-$ is the roll angular velocity change, $\varphi_{b\tau}^-=\varphi_b(t_\tau)$ and $\varphi_{b\tau}^+=\varphi_b(t_\tau^+)$ represent the roll angles before and after $\delta \tau_x$ is applied, respectively.

From~(\ref{Eq_Impact_Dynamics}) and~(\ref{Eq_Impulse_Effect}), it is clear that designing $\delta \tau_x$ (i.e., $\tau_\mathbb{I}$) is equivalent to identify the desired roll angular rate $\dot\varphi_{b\tau}^+$ and linear velocity $v_\tau^+$. We thus formulate the following optimization problem.
\vspace{-0mm}
\begin{subequations}\label{Eq_Optimal_Velocity}
\begin{align}
    \min_{\dot \varphi_{b\tau}^+,v_\tau^+}&\int_{t_\tau^+}^{t_\tau^++H_t} \bm{x}_e^T\bm{P}\bm{x}_e + {\bm u}^T \bm{Q}{\bm u}dt, \label{Eq_Optimal_Velocity_obj}\\
    \text{Subject to:}\; &(\ref{Eq_Bikebot_Kine}),~(\ref{Eq_Bikebot_Roll_2}),~(\ref{Eq_EIC_Control})\;\text{and}~(\ref{Eq_Impulse_Effect}), \label{Eq_Optimal_Velocity_EOM}\\
    & |\delta \tau_x| \leq \delta\tau_x^{\max}, |v_\tau^+| \leq v^{\max},  |\phi| \leq \phi^{\max}. \label{Eq_Optimal_Velocity_limit}
  \end{align}
\end{subequations}
\hspace{-0.0mm}where $\bm x_e = [\bm e_r^T\; \dot{\bm e}_r^T\;  e_b\; \dot e_b]^T$ is error vector, matrices $\bm P \in \mathbb{R}^{6\times 6}, \bm Q \in \mathbb{R}^{2\times 2}$ are symmetric positive definite, and $H_t>0$ is the predictive horizon. Condition~(\ref{Eq_Optimal_Velocity_obj}) denotes the tracking and balance errors and control efforts over the horizon $H_t$ and~(\ref{Eq_Optimal_Velocity_limit}) includes the constraints. Note that input $\bs{u}$ from~(\ref{Eq_EIC_Control}) is function of velocity $v$ and steering angle $\phi$. The optimization problem~(\ref{Eq_Optimal_Velocity}) can be interpreted as to search for the optimal conditions (i.e., velocities) to re-initialize the EIC-based control~(\ref{Eq_EIC_Control}). We solve~(\ref{Eq_Optimal_Velocity}) using the sequential quadratic programming (SQP) algorithm~\cite{nocedal2006sequential}.

The value of $\kappa$ is selected as small as possible to keep leg-terrain contact duration short. We use constant $\delta \tau_x$ during $\kappa$ and from~(\ref{Eq_Impulse_Effect}) we have
\begin{equation}
\label{Eq_Designed_Impulsive_Troque}
\delta \tau_x = \frac{1}{\kappa}J_t\left(\dot \varphi_{b\tau}^+-\dot \varphi_{b\tau}^-\right)=\frac{1}{\kappa}J_t \Delta \dot{\varphi}_b,\; t\in[t_\tau, t_\tau^+).
\end{equation}

To generate $\delta \tau_x $ given in~(\ref{Eq_Designed_Impulsive_Troque}), we need to design the leg-terrain interaction force $\bm F$ and the foot contact point $\bm r_i$. When leg touches on the terrain, contact point is located at $\bm r_i =[r_{ix}\; r_{iy}\; 0]^T$ and $\delta \tau_x $ is then expressed as $\delta \tau_x=r_{iy} F_{z}$, where $F_{z}=\bm F \cdot \bs{e}_z$, ${\bm e}_z=[0\;0\;1]^T$. Meanwhile, to eliminate the possible effects by $\delta \tau_y$ and  $\delta \tau_z $, we enforce additional constraints $F_{x}=F_{y}=0$ and $r_{ix}=0$ in the design. In practice, the wheel-obstacle impact often constrains and prevents the bikebot platform to rotate in the $y$- and $z$-axis directions and therefore, the magnitudes of $\delta \tau_y$ and  $\delta \tau_z $ are small and can be negligible. Given $r_{iy}$, the force $F_z$ then is determined as
\begin{equation}\label{Eq_Arm_Impulse}
  F_z=\frac{\delta \tau_x}{r_{iy}},\; F_z\leq F_z^{\max}
\end{equation}
where $F_z^{\max}$ is the maximum vertical ground reaction force. The leg-terrain contact point position $r_{iy}$ is offline tuned and adjusted in the experiment. The joint angles $\bs{\theta}$ is calculated by inverse kinematics~(\ref{Eq_Leg_Kinamatics}) and the joint torques are obtained as $\bm \tau_{\theta}=\bm{J}_{\theta}^{T}\left(\bm{R}_{\mathcal{H}}^{\mathcal{B}}\right)^T \bm{F}$ with $\theta_j^{\min} \leq \theta_{i,j} \leq \theta^{\max}_{j}$, $|{\tau}_{\theta_{i,j}}| \leq \tau_\theta^{\max}$, $i=L,R$, $j=1,2,3$.

\vspace{-1mm}
\subsection{Controller Integration and Analysis}
\label{Sec_Control_Integration}

\begin{figure}[h!]
\vspace{-1mm}
	\centering
	\includegraphics[width=3.5in]{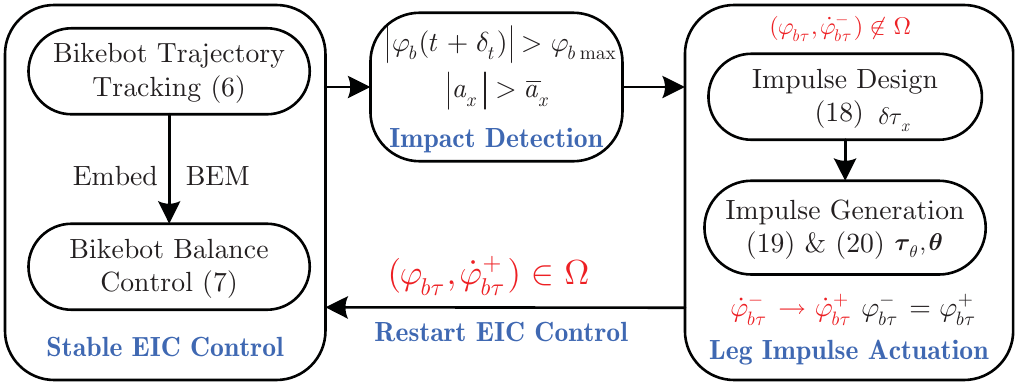}
	\vspace{-6mm}
	\caption{Finite state machines for the integrated EIC-based velocity and steering control and assistive impulse control.}
	\label{Fig_Control_Integration}
	\vspace{-3mm}
\end{figure}

The EIC-based control is continuous, while the impulse control has a discrete feature. Integration of those two control algorithms is thus a hybrid control strategy. Fig.~\ref{Fig_Control_Integration} illustrates the integration design of the two controllers.

The leg impulse control is triggered when the wheel hits the obstacle and the roll angle then reaches $\varphi_b^{\max}$, the maximum controllable roll angle under the EIC-based control. The real-time wheel-terrain impact is detected through the acceleration measurement (from IMU) $|a_x|$ greater than a pre-defined threshold $a_x^{\max}$. The model~(\ref{Eq_LSTM_Impact}) is used to estimate the state $\dot{\bm q}_f^*$ right after wheel-terrain impact. The roll motion is then predicted by~(\ref{Eq_Bikebot_Roll_3}) for a horizon $\delta_t$ at $t$ under the EIC-based control. If $\varphi_b(t)> \varphi_b^{\max}$ in $[t, t+\delta_t)$, the leg-terrain impulsive torque is initiated. In this case,~(\ref{Eq_Optimal_Velocity}) and~(\ref{Eq_Designed_Impulsive_Troque}) are used to design the impulsive torque $\delta \tau_x$ and~(\ref{Eq_Arm_Impulse}) is used to obtain the leg poses $\bs{\theta}$ and torques $\bs{\tau}_\theta$. Since the leg impulse control only helps balance task, the trajectory tracking still needs to be maintained by the EIC-based control. Thus the control design in~(\ref{Eq_EIC_Control}) is used to guarantee the trajectory tracking performance.

We analyze the bikebot roll motion under the impulse control. The roll dynamics for $t \in [t_\tau, t_\tau^+)$ is obtained as
\begin{equation}
  \dot{\varphi}_{b}(t)=\dot{\varphi}_{b\tau}^{-}+\frac{1}{J_{t}} \int_{t_\tau}^{t} \delta \tau_x d t=\dot{\varphi}_{b\tau}^{-}+\frac{t-t_{\tau}}{\kappa} \Delta \dot{\varphi}_{b}.
\label{Eq_Roll_Dynamics_Impulse}
\end{equation}
It is clear that $\dot{\varphi}_{b}(t_\tau^+)=\dot{\varphi}_{b\tau}^{+}$ and integrating~(\ref{Eq_Roll_Dynamics_Impulse}) leads to
\begin{equation}
\varphi_{b}(t)=\varphi_{b}(t_\tau)+\frac{\Delta \dot{\varphi}_b}{2\kappa}(t^2-t_\tau^2-2\kappa t_\tau).
\label{Eq_Roll_After_Impulse}
\end{equation}
Because of $\kappa \ll 1$, from~(\ref{Eq_Roll_After_Impulse}) we obtain that $\varphi_{b}(t_\tau^+)=\varphi_{b\tau}^+ \approx \varphi_{b}(t_{\tau})=\varphi_{b\tau}^-$, that is, $\delta \tau_x$ in~(\ref{Eq_Impulse_Effect}) is nearly an ideal impulse.

For roll motion at $t \geq t_\tau^+$, we simplify the steering torque $\tau_s= -{m_b}{h_G}\frac{{v^2}}{l}\c_\epsilon \tan \phi$ and~(\ref{Eq_Bikebot_Roll_3}) is linearized as
\begin{equation}
    \ddot{\varphi}_{b}-k_{1}^{2} \varphi_{b}+{k_{2}} \tan \phi=0,
\label{equ00}
\end{equation}
where $k_{1}^{2}=\tfrac{m_{b} h_{G}g}{J_{t}}$ and $k_2=\tfrac{m_{b} h_{G} v^{2}}{J_tl}$. Letting $\varsigma=t-t_\tau^+$ and $k_{3}=k_2/k_1$, the solution to~(\ref{equ00}) is then
\begin{equation*}
    \varphi_{b}(\varsigma)=c_1(\varsigma) \varphi_{b\tau}^++c_2(\varsigma) \dot{\varphi}_{b\tau}^+ +k_{3} \int_{0}^{\varsigma} c_{2}(\varsigma-s) \tan \phi(s) ds
\end{equation*}
where  $c_{1}(\varsigma)=\cosh(k_1\varsigma)$, $c_{2}(\varsigma)=\sinh(k_1\varsigma)$. Substituting~(\ref{Eq_Designed_Impulsive_Troque}) into the above equation, we obtain
\begin{align*}
 \varphi_{b}(\varsigma)=&c_{1}(\varsigma) \varphi_{b\tau}^-+c_{2}(\varsigma) \dot{\varphi}_{b\tau}^{-}+{k_{3}} \int_{0}^{\varsigma} c_{2}(\varsigma-s) \tan \phi(s) d s\\
 &+c_{2}(\varsigma) \kappa J_{t} \delta \tau_x.
\end{align*}
When state $(\varphi_{b\tau}^-,\dot{\varphi}_{b\tau}^{-})$ moves out of the region of attraction $\bs{\Omega}$ of the EIC-based control, that is, $(\varphi_{b\tau}^-,\dot{\varphi}_{b\tau}^{-})\notin \bs{\Omega}$, they have the same sign; see Fig.~\ref{Fig_DOA}. In this case, steering angle $\phi(t)$ needs to reach maximum value $\phi_{\max}$ for balance control and we then approximate roll angle as
\begin{equation*}
{\varphi _b}(\varsigma) \approx {c_1}(\varsigma){\varphi_{b\tau}^-} + {c_2}(\varsigma)\dot \varphi_{b\tau}^-  + {k_3}{c_1}(\varsigma)\tan {\phi _{\max }} + {c_2}(\varsigma)\kappa {J_t}\delta \tau_x.
\end{equation*}
We collect all terms related to $e^{k_1\varsigma}$ in the above equation that might cause unstable motion and to prevent the divergence, we obtain the necessary condition
\begin{equation}
\text{\hspace{-3mm}} \sign(\dot{\varphi}_{b\tau}^-)\left(\varphi_{b\tau}^-+\dot{\varphi}_{b\tau}^{-}+k_{3} \tan \phi_{\max }+\kappa J_{t} \delta \tau_x\right)<0.
\label{equ01}
\end{equation}

If $\dot{\varphi}_{b\tau}^->0$, the applied impulsive torque $\delta \tau_x$ is negative and is implemented by the right leg actuation. Similarly, if $\dot{\varphi}_{b\tau}^-<0$, $\delta \tau_x$ is positive and the left leg is used in actuation. From~(\ref{equ01}) we obtain the condition for $\delta \tau_x$ as
\begin{equation}
|\delta \tau_{x}| > \delta \tau^{\min}_{x}:=\frac{|\varphi_{b \tau}^{-}+\dot{\varphi}_{b \tau}^{-}+k_{3} \tan \phi_{\max}|}{\kappa J_{t}}.
\label{equ02}
\end{equation}
Condition~(\ref{equ02}) implies that under the EIC control~(\ref{Eq_EIC_Control}), when the bikebot is about to lose balance (i.e., $(\varphi_{b\tau}^-, \dot\varphi_{b\tau}^-) \notin \bs{\Omega}$), the impulsive torque magnitude $|\bs{\delta} \tau|$ should be greater than $\delta \tau^{\min}_{x}$ to generate $(\varphi_{b\tau}^+, \dot\varphi_{b\tau}^+) \in \bs{\Omega}$ to recovery to balance state.

The stability of the hybrid control strategy can be shown by two steps. First, the bikebot trajectory tracking under the EIC control design is exponentially stable and the tracking errors converge to a small ball around zero~\cite{GetzPhD}. Secondly, the impulse input provides the correction effect and moves the roll motion to the domain of attraction of EIC control. With $(\varphi_{b\tau}^+, \dot\varphi_{b\tau}^+)\in \bs{\Omega}$, the stable trajectory tracking under EIC control will restart. The stability of the system then is guaranteed in a piecewise style and we can view the impulse input as a re-initialization effect. We will show this in experiments in the next section.

\renewcommand{\arraystretch}{1.3}
\setlength{\tabcolsep}{0.04in}
\begin{table*}[htb!]
	\centering
	\caption{Physical parameters of the bikebot and the assistive legs}
	\vspace{-1mm}
	\begin{tabular}{|c|c|c|c|c|c|c|c|c|c|c|c|}
	   \hline\hline
	   $m_b$ (kg) & $J_b$ (kgm$^2$) & $J_z$ (kgm$^2$) & $l$ (m) & $R_w$ (m) & $l_G/h_G$ (m) & $l_b/h_b$ (m) & $\varepsilon$ (deg) & $l_j$ (m)& $l_x, l_y$ (m) & $\tau_{\theta_j}^{\max}$ (Nm) & $\theta^{\max}_j/\theta^{\min}_j$ (deg)\\
	   \hline
	   $24$ & $0.25$ & $0.5$ & $0.87$ & $0.225$&$-0.04/0.35$ & $-0.14/0.26$ & $17$  & $0.075/0.212/0.207$ & $0.1/0.13$ & $25$ & $150/-150$\\
		\hline\hline
	\end{tabular}
	\label{Table_Bikebot_Parameters}
\vspace{-0mm}
\end{table*}

\section{Experiments}
\label{Sec_Experiment}

\subsection{Experiment Setup}
\begin{figure}[h!]
	\vspace{-1mm}
	\centering \includegraphics[width=2.8in]{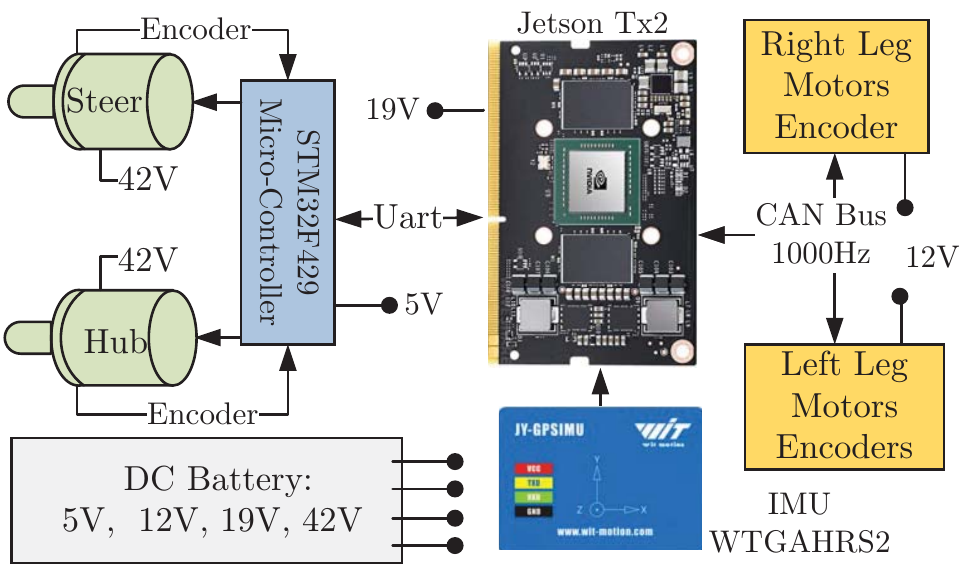}
	\vspace{-2mm}
	\caption{Embedded system design schematic for the bikebot.}
	\label{Fig_Bikeleg_Control_Board}
	\vspace{-1mm}
\end{figure}

\begin{figure}[h!]
	\vspace{-2mm}
	\centering
	\includegraphics[width=3.4in]{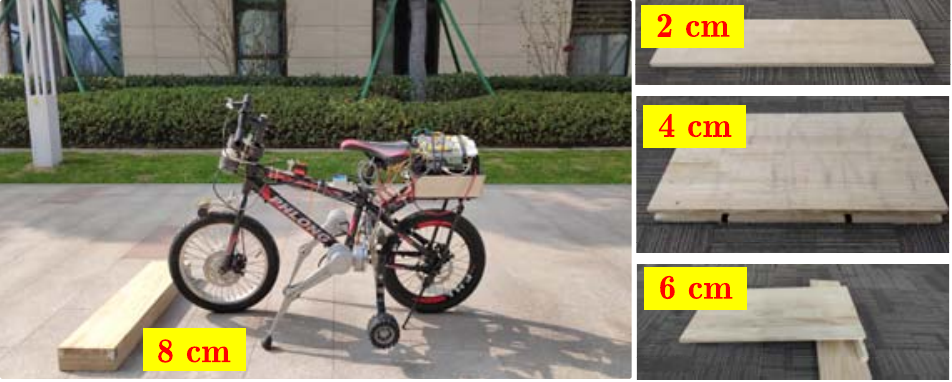}
	\vspace{-1mm}
	\caption{Bikebot experiments for various obstacle crossing tests. Wooden blocks with different heights are used as obstacles in experiments.}
	\label{Fig_Setup}
	\vspace{-0mm}
\end{figure}

\setcounter{figure}{5}
\begin{figure*}[ht]
	\hspace{-4mm}
	\subfigure[]{
		\label{Fig_Force_Validate}
		\includegraphics[width=2.25in]{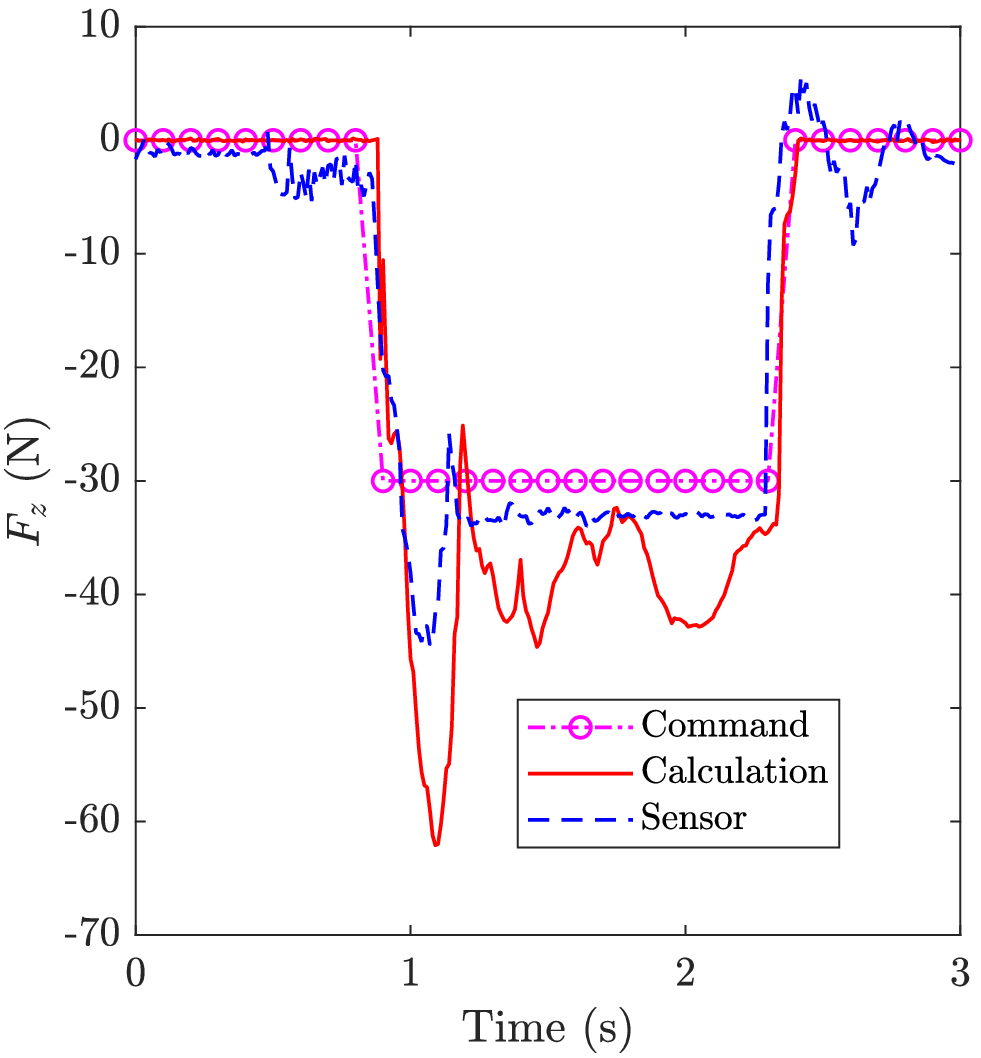}}
	\hspace{-2mm}
	\subfigure[]{
		\label{Fig_Prediction}
		\includegraphics[width=2.4in]{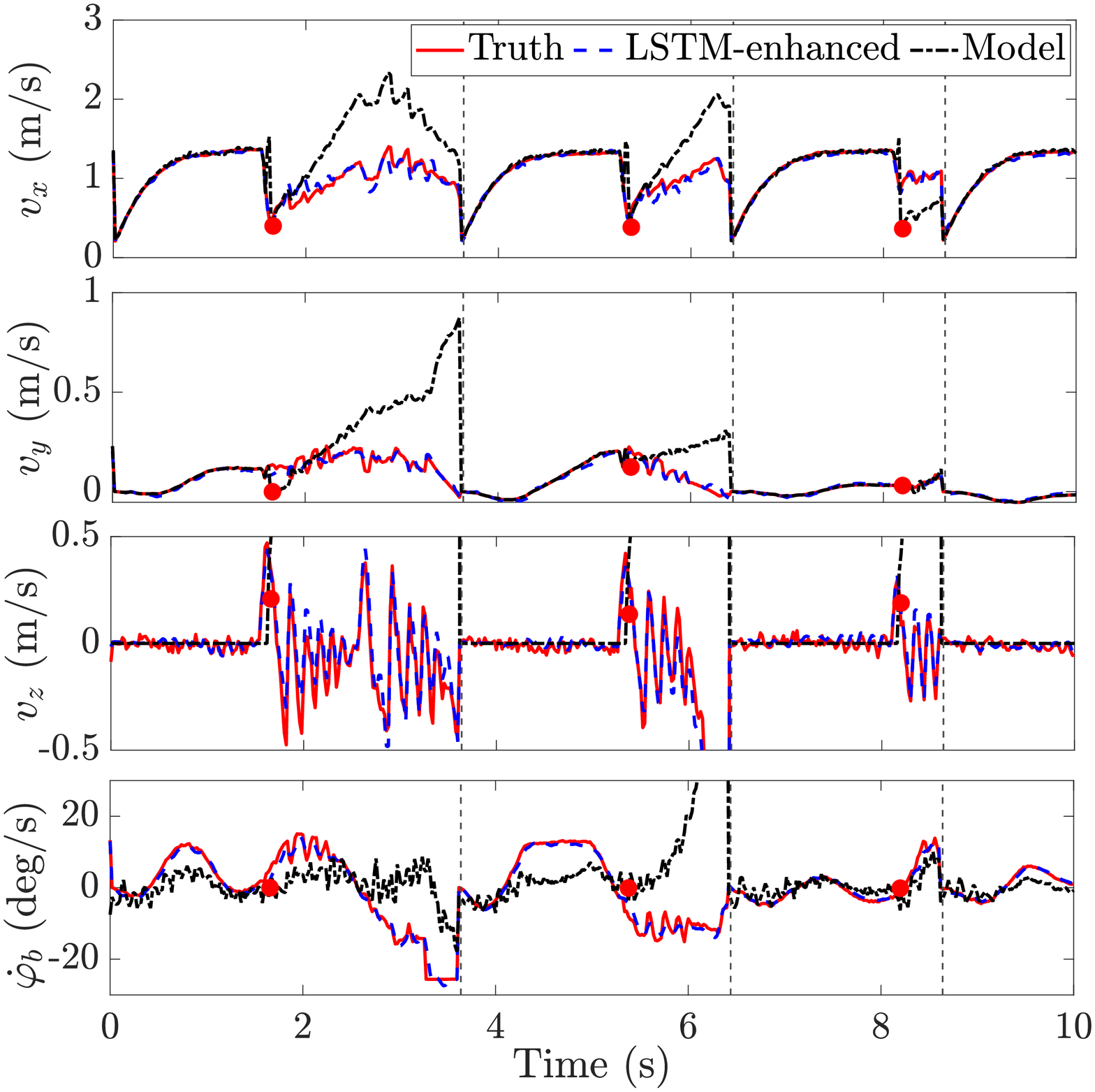}}
	\hspace{-3mm}
	\subfigure[]{
		\label{Fig_PredError}
		\includegraphics[width=2.36in]{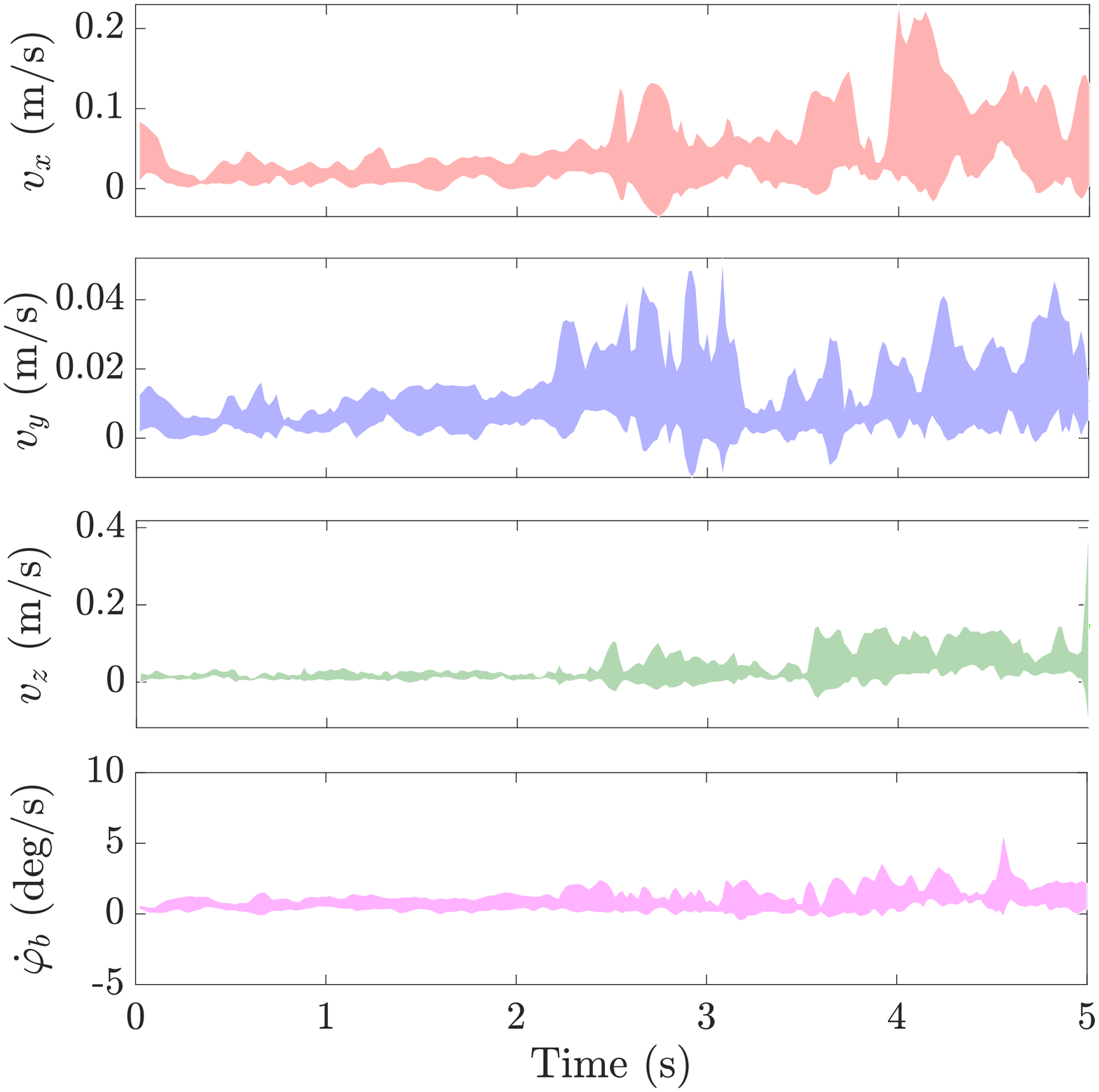}}
	\vspace{-1mm}
	\caption{(a) Verification experiment of the leg-terrain interaction force control by using the join torque feedback. (b) Wheel-obstacle impact model estimation results by the impact model and the LSTM-based enhancement. The red circular markers "{\color{red} $\bullet$}" denote the $\bs{q}_f^+$ estimates by the impact model. The vertical lines indicates different experiment trials. (c) Prediction errors of multiple trials with mean and standard deviation. The time duration is normalized to 5~s. }
	\label{Fig_Impact_Prediction}
	\vspace{-3mm}
\end{figure*}

\begin{figure*}
	\centering
	\subfigure[]{
		\label{Fig_Impulse_Control_r}
		\includegraphics[width=2.27in]{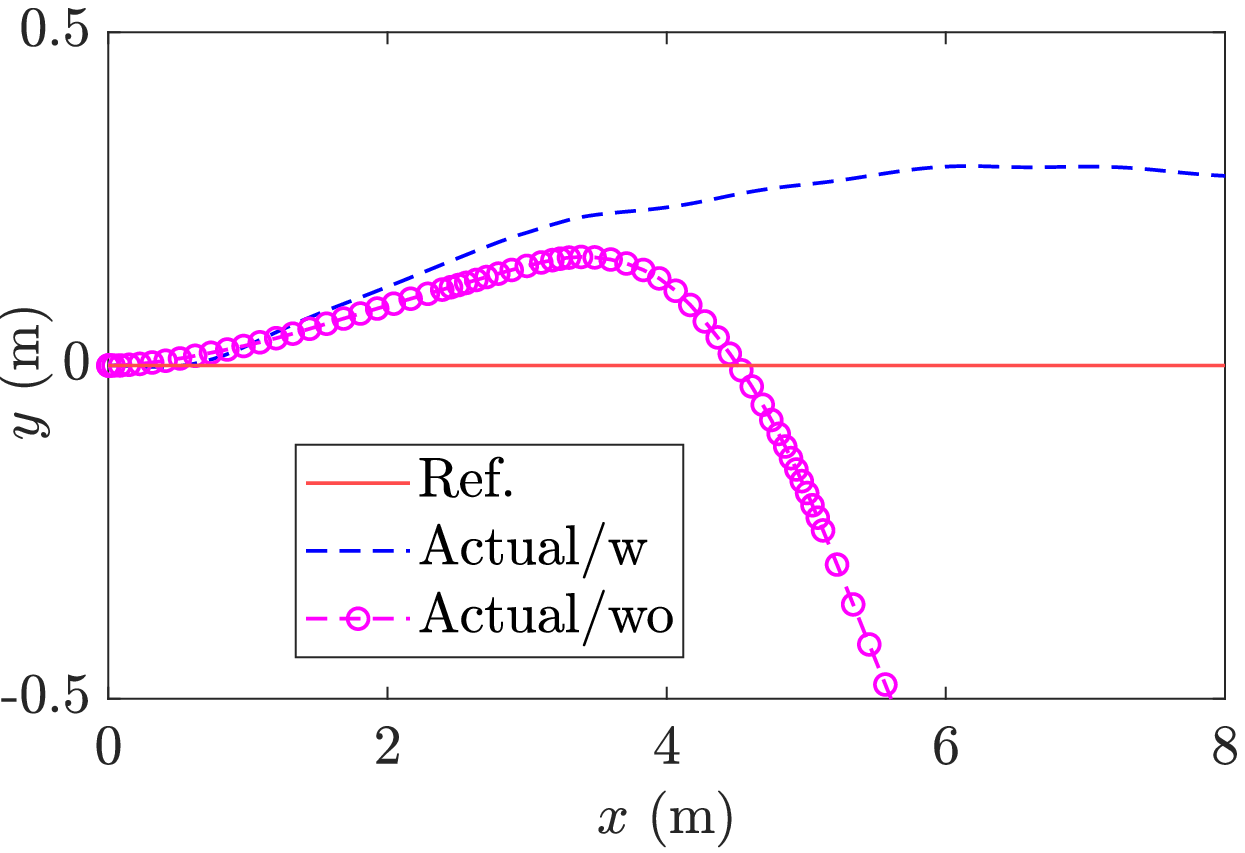}}
	\hspace{-2mm}\vspace{-2mm}
	\subfigure[]{
		\label{Fig_Impulse_Control_v}
		\includegraphics[width=2.25in]{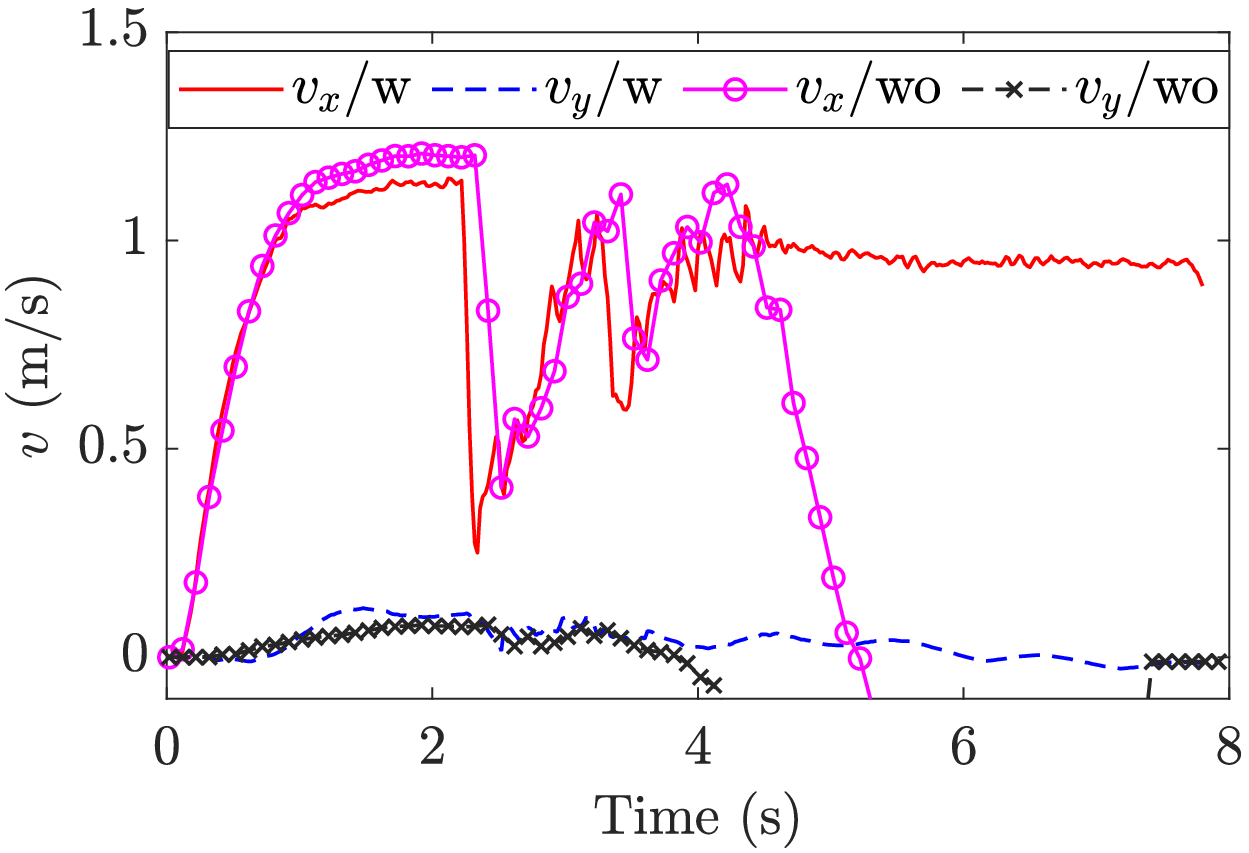}}
	\hspace{-2mm}
	\subfigure[]{
		\label{Fig_Impulse_Control_phi_b}
		\includegraphics[width=2.25in]{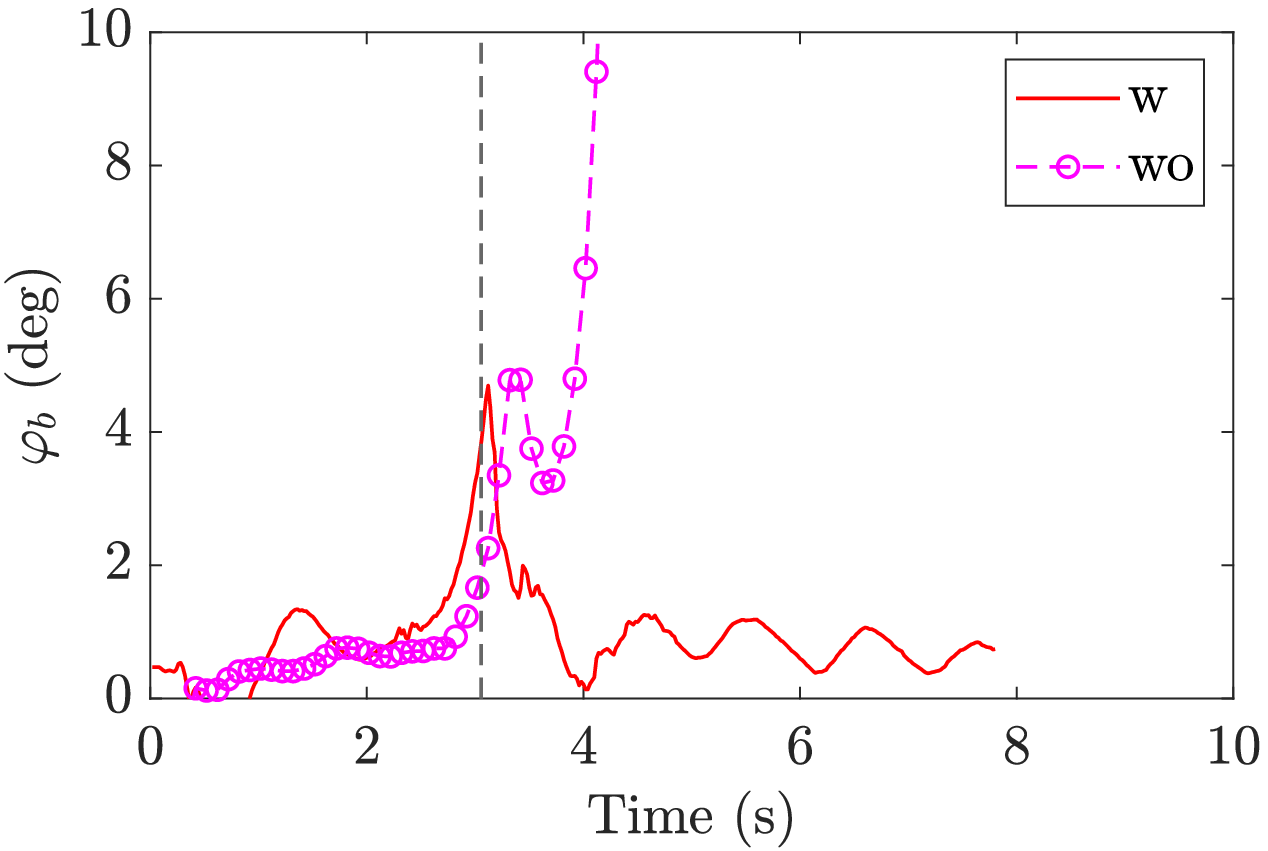}}
	\hspace{-2mm}
	\subfigure[]{
		\label{Fig_Impulse_Control_dot_phi_b}
		\includegraphics[width=2.29in]{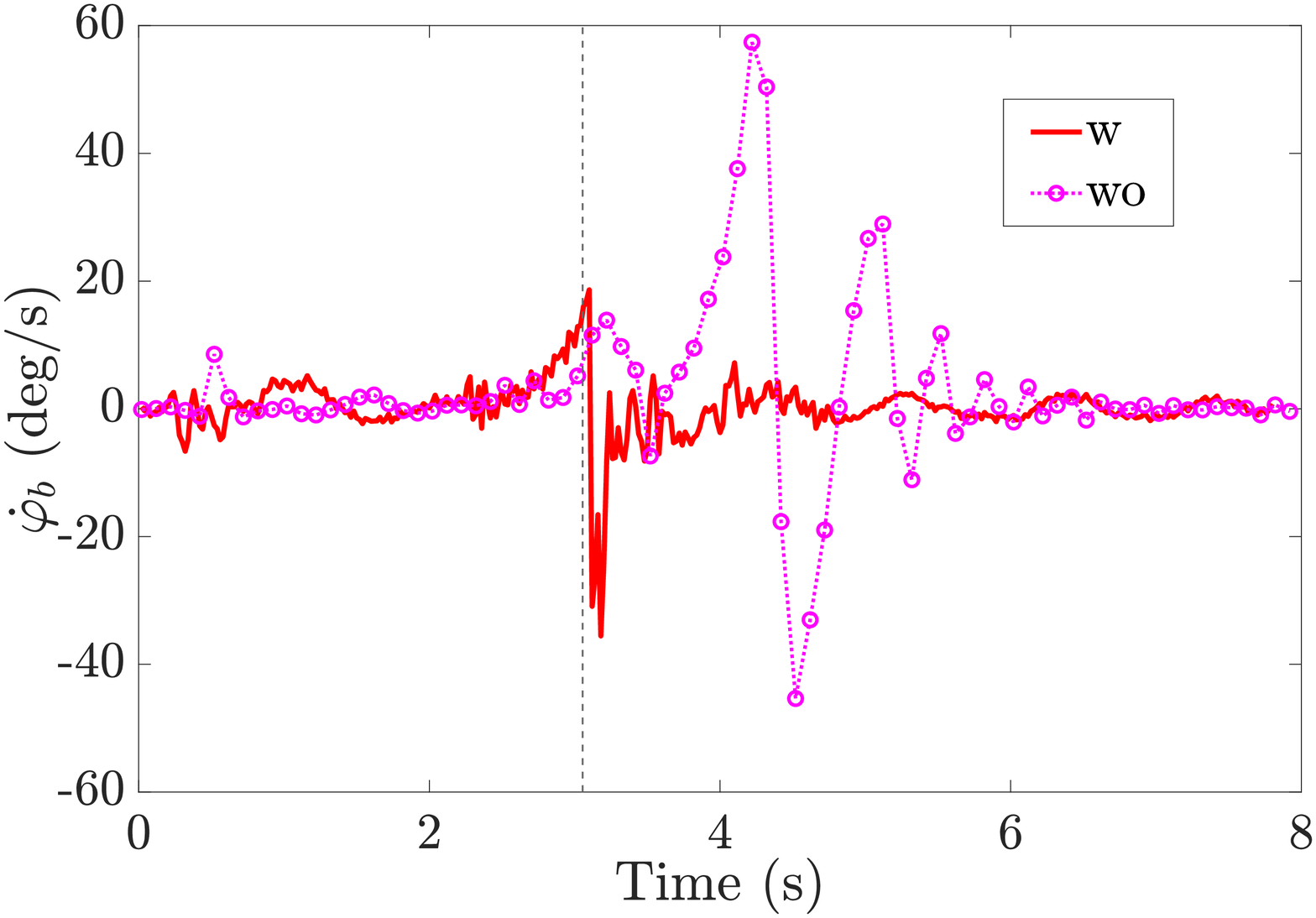}}
	\hspace{-2mm}
	\subfigure[]{
		\label{Fig_Impulse_Control_Leg_2}
		\includegraphics[width=2.25in]{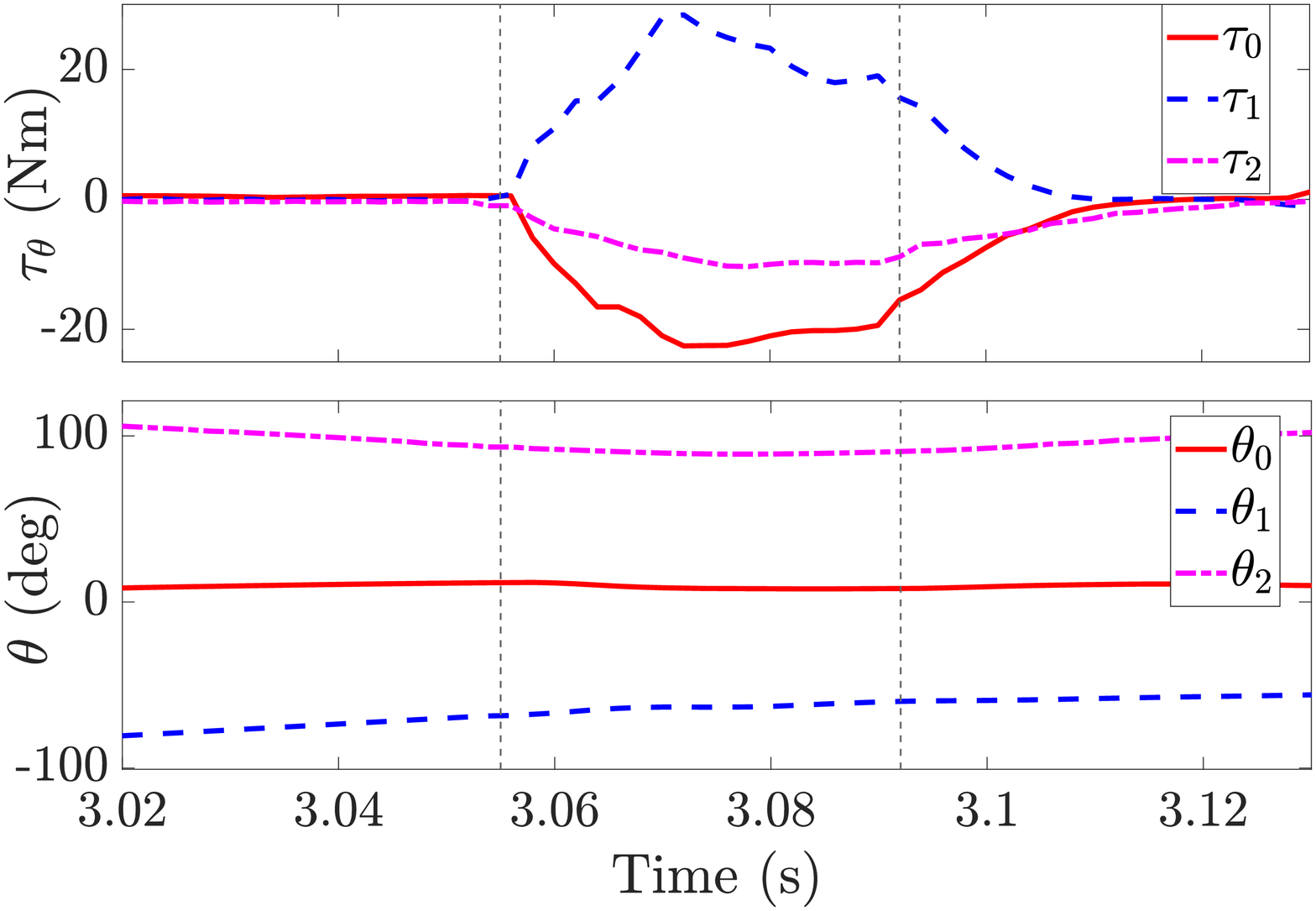}}
	\hspace{-2mm}
	\subfigure[]{
		\label{Fig_Impulse_Control_Leg_1}
		\includegraphics[width=2.28in]{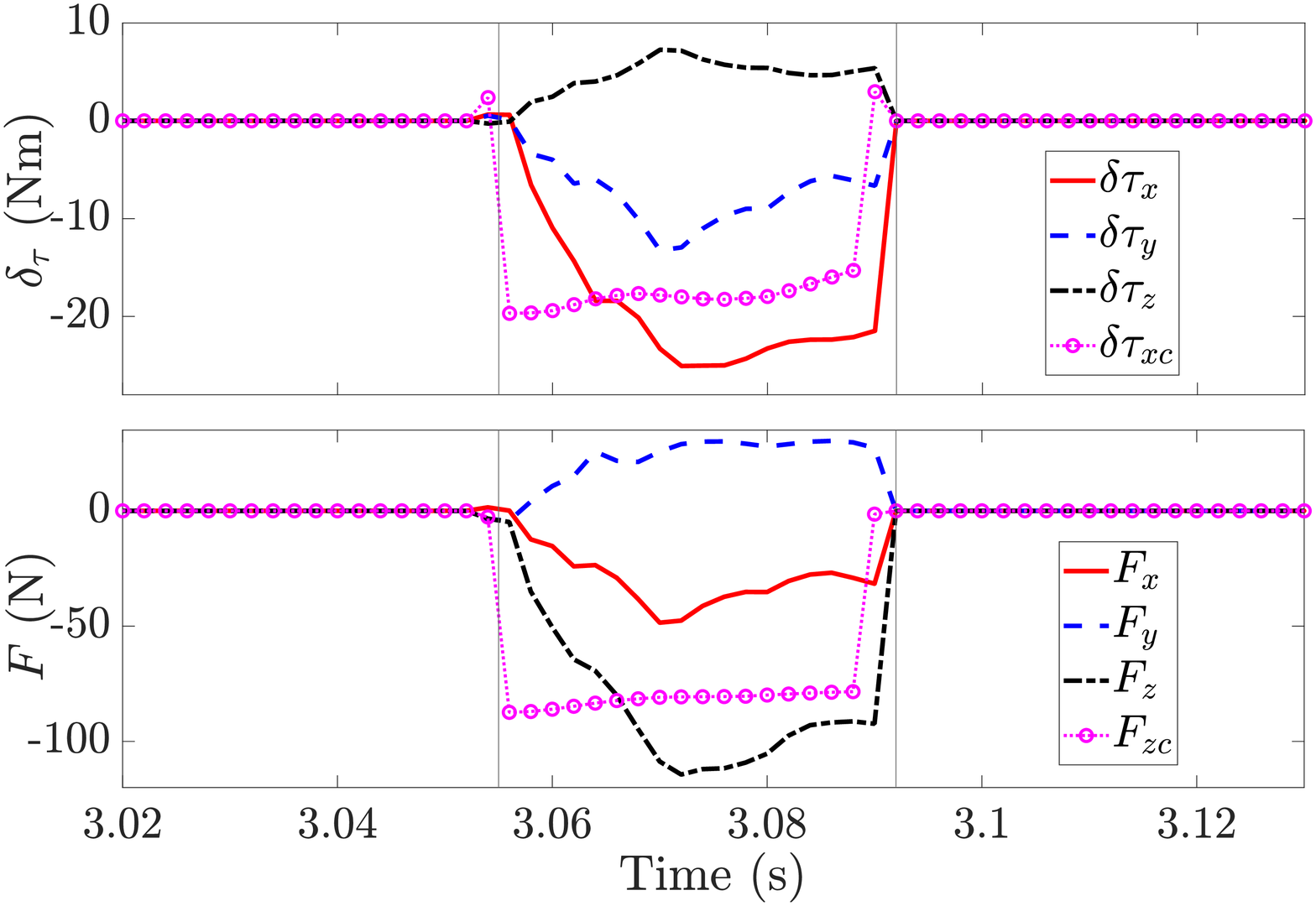}}
	\vspace{-2mm}
	\caption{Leg assistative impulse control for a straight line tracking with impulse time duration being $50$ ms. (a) Bikebot position, (b) bikebot velocity, (2) bikebot roll angle, (d) bikebot roll angular velocity, (e) leg joint torque and joint angle, (f) leg-terrain interaction force and torque. The vertical lines in (c)-(f) indicate impulse control time. }
	\label{Fig_Impulse_Control}
	\vspace{-2mm}
\end{figure*}

Fig.~\ref{Fig_BikePhoto} shows the bikebot prototype and Fig.~\ref{Fig_Bikeleg_Control_Board} illustrates the embedded system design. The bikebot velocity was measured by an encoder mounted and an IMU was mounted on the frame to measure the roll and yaw motion. The onboard embedded system includes a low-level real-time microcontroller and the upper-level ROS-based platform (Nvidia Jetson TX2). The assistive leg control was implemented on the ROS-based platform through a CAN bus interface. The leg control and the EIC-based control were implemented at frequencies of $1000$ and $50$~Hz, respectively. No force/torque sensor was used to measure the leg-terrain interaction force and instead, the leg joint torques were measured in real time to estimate the interaction forces for impulse control. We conducted experiments and validated such an approach by using an external 6-DOF force sensor (Model M3703C from Sunrise Instruments, Nanning, China). Fig.~\ref{Fig_Setup} shows the experiment setup. We used a set of wooden blocks to construct obstacles with different heights. Irregular obstacles were also constructed using these wooden blocks in experiments as shown in the figure.

Table~\ref{Table_Bikebot_Parameters} lists the values of the physical model parameters and the leg joint angle and torque bounds. In experiments, the impulsive torque duration was selected as $\kappa = 50$~ms and the prediction horizon in~(\ref{Eq_Optimal_Velocity}) was chosen $H_t = 1$ s (i.e., 50 control cycles). Other optimization and controller parameters values include: $\delta_t=200$~ms, $\delta \tau_x^{\max}=30$~Nm, $F_z^{\max}=190$~N, $\phi^{\max} = 30$~degs, $v^{\max} = 1.5$~m/s, $a_x^{\max}=5$~m/s$^2$, $\varphi_b^{\max}=5$ degs, $\bs P=\diag(1,1,1,1,10,10)$, and $\bs Q=\diag(10,10)$. The EIC-based control parameters were chosen as $a_1=3,a_2=6,a_3=10,b_1=180$, and $b_2=25$. The parameters selected above are first tested through simulations and preliminary experiment. The leg force limit is obtained and validated through a static test similar as Fig.~\ref{Fig_Force_Validate}.

\subsection{Experimental Result}

Fig.~\ref{Fig_Force_Validate} shows the leg force control results. A leg-terrain interaction force $F_z=30$ N with duration $1.5$~s was commanded. We compared the force using the joint torque feedback and analytical calculation ($\bs{F}=(\bm{J}^T_\theta\left(\bm R_\mathcal{H}^\mathcal{B}\right)^T)^{-1}\bm \tau_\theta$) with the ground truth by the external force sensor. It is clear that the joint torque feedback can be used to maintain the force control as the results match the measurements. Fig.~\ref{Fig_Prediction} further shows the impact model prediction results. The LSTM was trained with over $M=10000$ data sets and $N=10$. These data were collected in experiments to run across obstacles with various heights. The ground truth information of the bikebot motion was obtained from an optical motion capture system (from Vicon Inc.) From the results, the roll and velocity estimations by the impact model were not in good agreement with the ground truth, while with the LSTM-based enhancement, the results closely matched the ground truth. Fig.~\ref{Fig_PredError} further shows the prediction errors by the LSTM-enhanced results from multiple experiments, which consistently confirm small errors. The LSTM prediction running time was around $16$~ms on average and can be implemented in real time.

\setcounter{figure}{7}

\begin{figure*}[htp!]
	\centering
    \hspace{-4mm}
\subfigure[]{
	\label{Fig_Multiple_Trial_A}
	\includegraphics[width=2.45in]{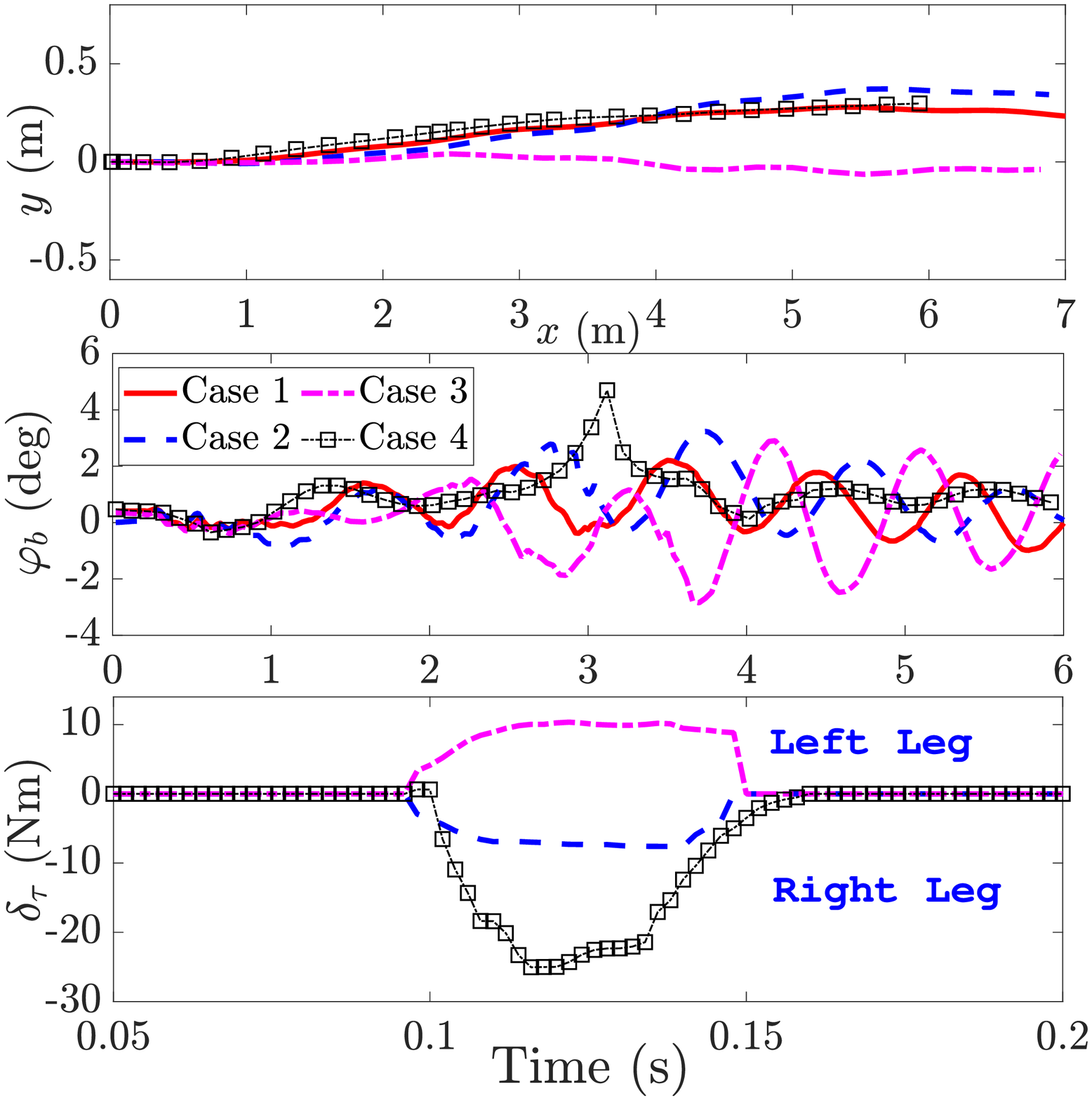}}
    \hspace{-3mm}
	\subfigure[]{
		\label{Fig_Multiple_Impulse_A}
		\includegraphics[width=2.35in]{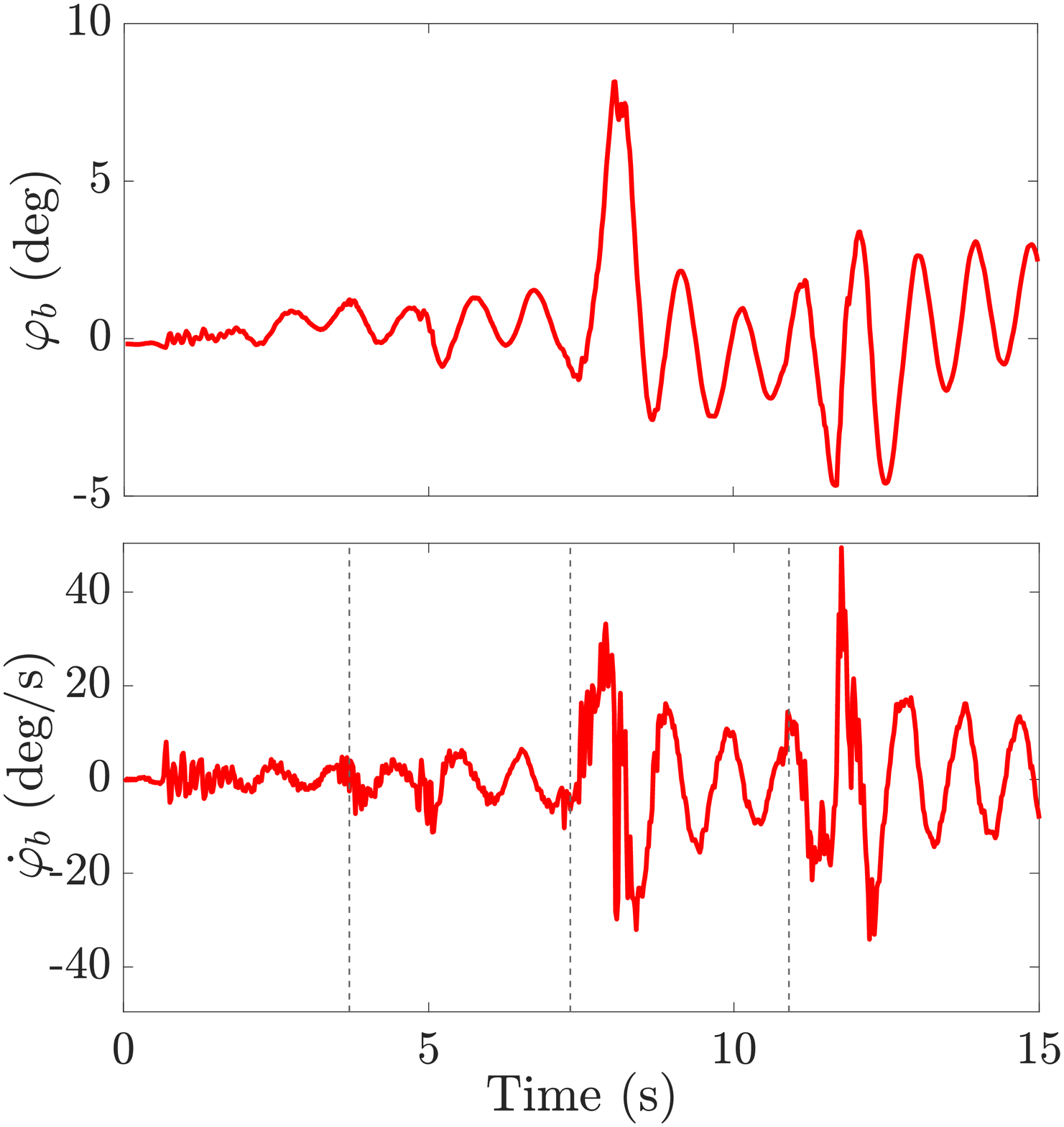}}
    \hspace{-3mm}
	\subfigure[]{
		\label{Fig_Multiple_Impulse_B}
		\includegraphics[width=2.32in]{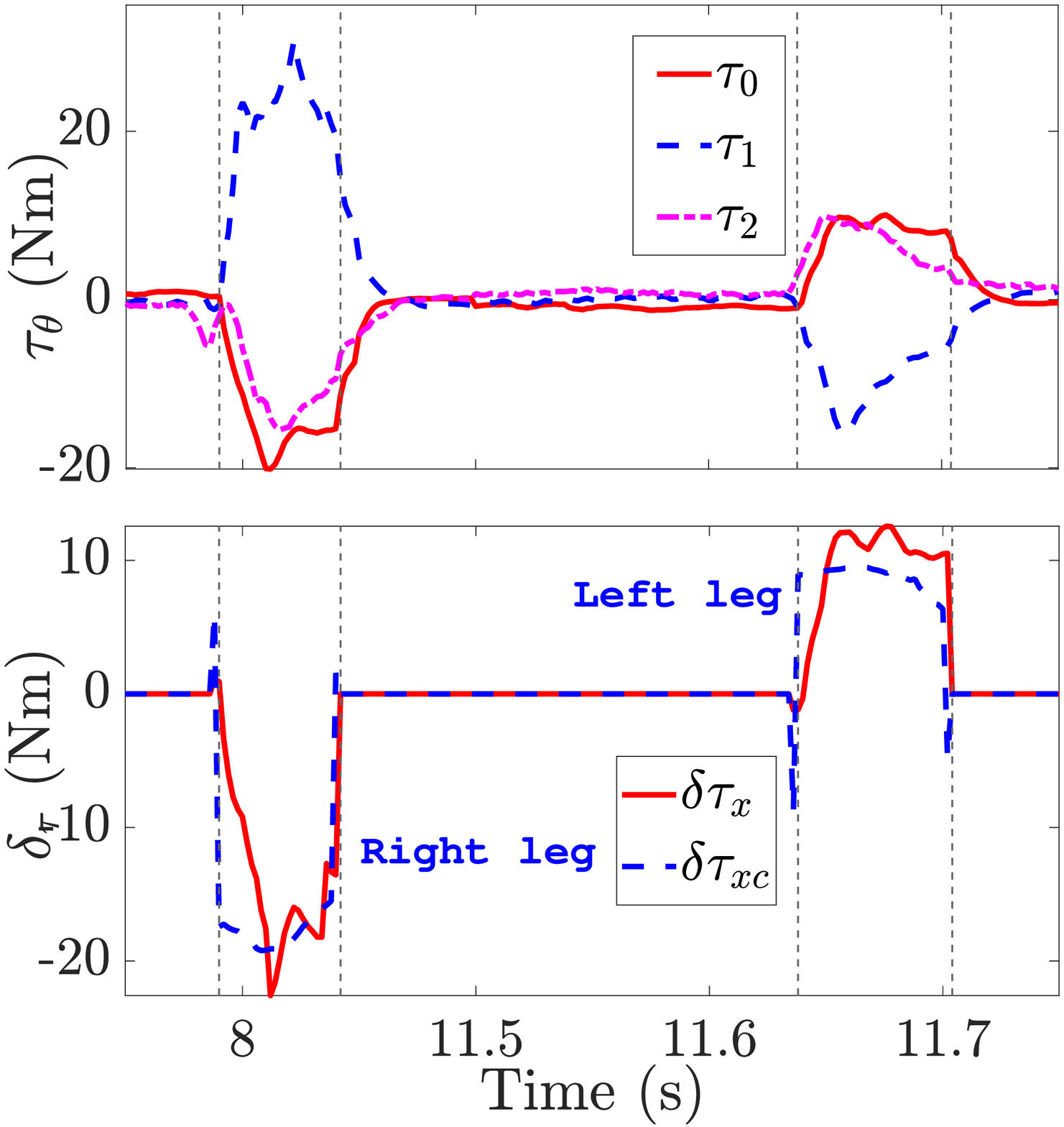}}
	\vspace{-1mm}
	\caption{(a) Experimental results for straight-line tracking and crossing one obstacle with different heights. Case 1: $h_o=2$ cm, Case 2: $h_o=4$ cm, Case 3: $h_o=5$ cm, Case 4: $h_o=6$ cm. Top: Planar trajectory; Middle: Roll angles; Bottom: Applied impulse torques.(b) and (c) are experimental results to run crossing three obstacles in one run at heights $h_o=3.5, 6, 5$ cm continuously. (b) Bikebot roll motion. The wheel-obstacle impacts happen at the moments indicated by the vertical lines. (c) Leg joint torques and impulse torques. The positive and negative $\delta \tau_x$ are applied by the left and right legs, respectively.}
	\label{Fig_Multiple_Impulse}
	\vspace{-2mm}
\end{figure*}

Fig.~\ref{Fig_Impulse_Control} shows the impulse control results in a straight-line tracking experiment. A rectangular wooden block with height $h_o=7.8$ cm was used as the obstacle. Figs.~\ref{Fig_Impulse_Control_r} and~\ref{Fig_Impulse_Control_v} show the bikebot trajectory and velocity with and without the impulse input. At $t=2.3$ s, the wheel hit the obstacle and the velocity significantly dropped about $1$~m/s. In Fig.~\ref{Fig_Impulse_Control_phi_b}, the roll angle increased monotonically to $\varphi_b^{\max}=5$~degs and only the steering control cannot recover the balance. When the impulse was triggered at around $t_\tau=3.05$~s, the right leg hit the ground to generate the impulsive torque $\delta \tau_x=20$~Nm; see Fig.~\ref{Fig_Impulse_Control_Leg_2}. The roll velocity changed from $\dot \varphi_{b\tau}^-=20$ to $\dot \varphi_{b\tau}^+=-40$ deg/s in Fig.~\ref{Fig_Impulse_Control_dot_phi_b}. Under this impulse control, the roll angle moved back to around zero and the bikebot motion was back into the region of attraction $\bs{\Omega}$ of the EIC-based control with the increased speed $v=1.1$~m/s at $t=4$~s. The result without the impulse input showed the loss the balance with the roll angle over $10$~degs at this moment. Figs.~\ref{Fig_Impulse_Control_Leg_2} and~\ref{Fig_Impulse_Control_Leg_1} show the right leg joint angles and torques, impulsive torques, and the leg-terrain interaction forces. Within the impulse duration, the roll angle change (around $0.3$~degs) was small. The foot contact angle $\theta_0 \approx 0$ implies the leg link touched the ground nearly vertically. Under this configuration, the vertical contact force $F_z$ played the major role in impulsive torque generation. These results confirmed the effectiveness of the impulsive torque control design.

We further demonstrate the integrated control of the bikebot running across obstacles with different heights. Fig.~\ref{Fig_Multiple_Trial_A} shows results of the bikebot crossing four obstacles separately at heights $h_o=2$ cm (Case 1), $4$ cm (Case 2), $5$ cm (Case 3) and $6$ cm (Case 4). For a thin obstacle (Case 1), the EIC-based control successfully maintained the balance without impulse input, showing the robustness of the control design. For others three cases, the wheel-obstacle impacts were significant due to high steps and the leg assistive control was needed. To run crossing obstacles with large heights, the roll velocity changes were large and significant impulse inputs were needed; see the bottom plot in Fig.~\ref{Fig_Multiple_Trial_A}. We further conducted experiments to run across three obstacles continuously with heights $h_o=3.5$, $6$, and $5$~cm. Figs.~\ref{Fig_Multiple_Impulse_A} and~\ref{Fig_Multiple_Impulse_B} show the roll motion and the leg assistive impulsive torques. The bikebot successfully ran across the first obstacle by the EIC-based control and for the last two obstacles, the impulse control was triggered to regain the balance. The roll angles moved back to around zero deg in a short duration and the left and right legs were sequentially used to generate the impulsive torques, as shown in Fig.~\ref{Fig_Multiple_Impulse_B}.

\begin{figure}[h!]
	\centering
	\includegraphics[width=3.4in]{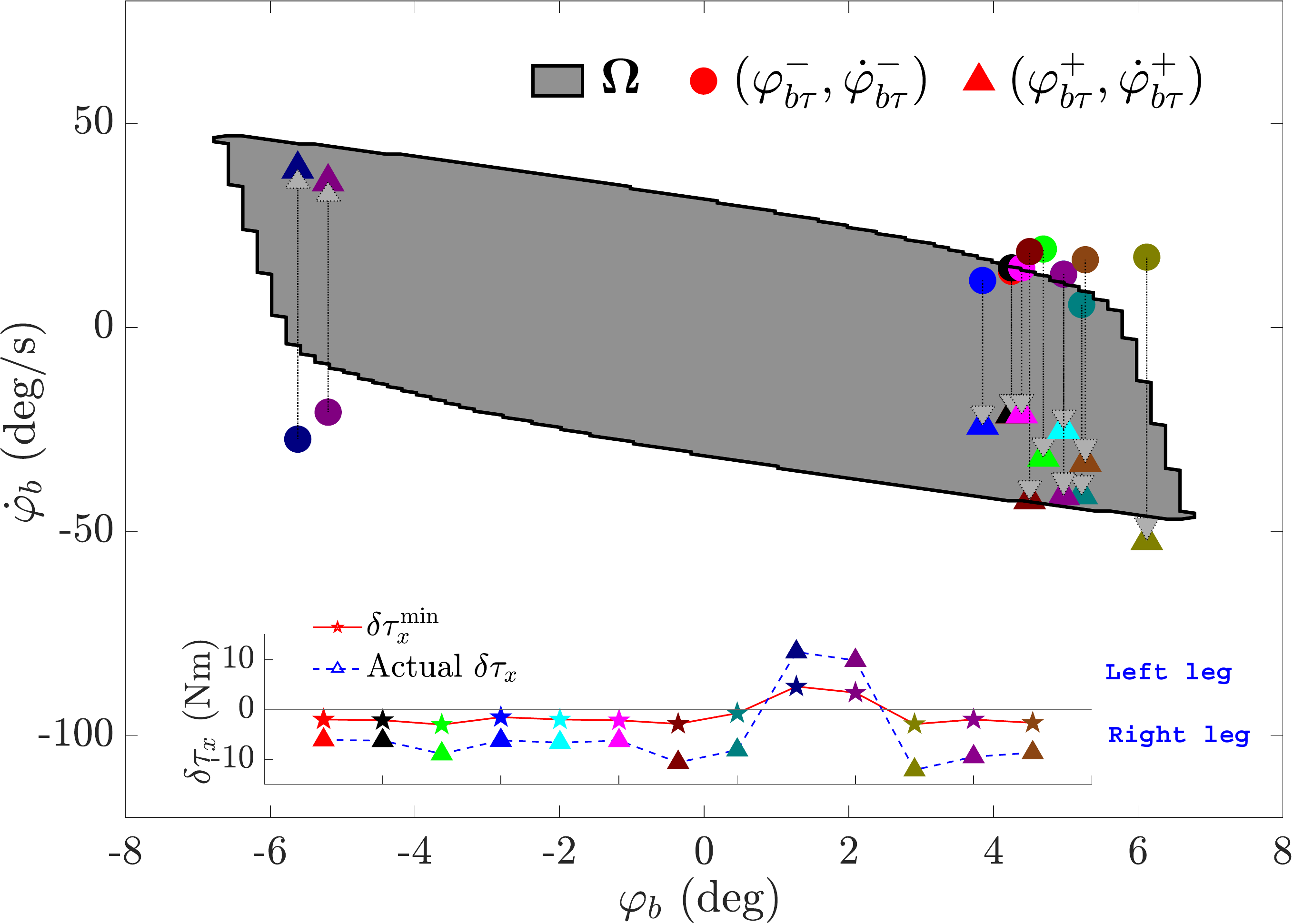}
	\vspace{-1mm}
	\caption{The phase portrait of the pre-impulse and post-impulse states and their relationship with the estimated region of attraction of the EIC-based control. The ``$\bullet$'' and ``$\blacktriangle$'' markers indicated the pre-impulse state $(\varphi_{b\tau}^-,\dot{\varphi}_{b\tau}^-)$ and post-impulse state $(\varphi_{b\tau}^+,\dot{\varphi}_{b\tau}^+)$ in experiments. }
	\label{Fig_DOA}
	\vspace{-2mm}
\end{figure}

Fig.~\ref{Fig_DOA} shows the relationship of the bikebot roll motion in the $(\varphi_{b},\dot\varphi_{b})$ phase plane under the integrated balance control. The shaded area shows the region of attraction $\bs{\Omega}$ under the EIC-based controller. We obtained $\bs{\Omega}$ by computation with imposing the same constraints in~(\ref{Eq_Optimal_Velocity}) for the straight-line tracking task. We plotted the bikebot roll motion before (circular makers) and after (triangular markers) the impulse control in multiple experimental runs. It is clear that the pre-impulse state $(\varphi_{b\tau}^-,\dot\varphi_{b\tau}^-)$ are at the boundary or outside of $\bs{\Omega}$, while the post-impulse state $(\varphi_{b\tau}^+,\dot\varphi_{b\tau}^+)$ jumped and moved back into $\bs{\Omega}$. This implies that the designed impulse control helps to bring the roll motion state back to the region of attraction and re-initialize the EIC-based control to achieve the tracking and balance tasks. The bottom plot in Fig.~\ref{Fig_DOA} further shows that the actual impulsive torques $\delta \tau_x$ in experiments satisfy the analytical bounds $\delta \tau_x^{\min}$ that are computed by~(\ref{equ02}). These results confirm the integrated impulse balance control design and the stability analysis.

\section{Conclusion}
\label{Sec_Conclusion}

This paper presented an integrated impulse control with the nonlinear trajectory tracking and balance (i.e., EIC-based control) for a single-track autonomous bikebot to safely run across obstacles. Two assistive legs were attached to the bikebot to provide the leg-terrain impulsive torques when crossing obstacles. The impulse control was built on an optimization-based approach and a switching strategy was designed to integrate with the EIC-based controller. An LSTM-based model was used to improve the estimation accuracy for the wheel-obstacle impact in real time. The experimental results demonstrated that with the proposed mechatronic and control design, the bikebot can navigate crossing step-like obstacles with heights more than one third of the wheel radius. Extensive experiments were presented to demonstrate the robustness of the control systems design over various obstacles.

\vspace{0mm}
\bibliographystyle{IEEEtran}
\bibliography{YiRef}

\end{document}